\documentclass[11pt]{article}

\usepackage[final]{acl}

\usepackage{latexsym}

\usepackage{fontspec}
\IfFontExistsTF{[fonts/texgyretermes-regular.otf]}{%
  \setmainfont{texgyretermes}[Path=fonts/, Extension=.otf,
    UprightFont=*-regular, BoldFont=*-bold, ItalicFont=*-italic, BoldItalicFont=*-bolditalic]%
}{%
  \setmainfont{TeX Gyre Termes}%
}
\IfFontExistsTF{[fonts/Norasi.otf]}{%
  \newfontfamily\thaifont{Norasi}[Path=fonts/, Extension=.otf, UprightFont=*, BoldFont=*-Bold]%
}{%
  \newfontfamily\thaifont{Norasi}%
}
\IfFontExistsTF{Inconsolatazi4-Regular.otf}{%
  \setmonofont{Inconsolatazi4}[Extension=.otf, UprightFont=*-Regular, BoldFont=*-Bold,
    Scale=MatchLowercase, WordSpace={1,1,1}]%
}{}

\usepackage{microtype}

\usepackage{graphicx}
\usepackage{amssymb}
\usepackage{booktabs}
\usepackage{amsmath}
\usepackage{tabularx}
\usepackage{placeins}
\usepackage{float}
\usepackage{listings}
\usepackage{needspace}
\usepackage{multirow}

\title{Reference-Grounded Data Curation for Instruction-Following Thai-English Machine Translation}

\author{%
  \textbf{Thodsaporn Chay-intr\textsuperscript{1,2}},
  \textbf{Krittapad Harnchang\textsuperscript{1}}\thanks{Work done while at iApp Technology.},
  \textbf{Mahannop Thabua\textsuperscript{1}}\footnotemark[1] \\
  \textbf{Kobkrit Viriyayudhakorn\textsuperscript{1,3}},
  \textbf{Thanaruk Theeramunkong\textsuperscript{2,4}} \\
  \textsuperscript{1}iApp Technology, Thailand \\
  \textsuperscript{2}Intelligent Informatics and Service Innovation Research Center, Thailand \\
  \textsuperscript{3}Artificial Intelligence Entrepreneur Association of Thailand (AIEAT), Thailand \\
  \textsuperscript{4}Sirindhorn International Institute of Technology, Thammasat University, Thailand \\
  \texttt{\{t.chayintr, krittapadpor12348, peemmygg\}@gmail.com} \\
  \texttt{kobkrit@aieat.or.th, thanaruk@siit.tu.ac.th}}

\begin{document}
\maketitle

\begin{abstract}

Instruction-following machine translation (IF-MT) requires respecting prompt-level rules on terminology, formatting, and register. Rule compliance typically trades off against translation quality, a tension that general-purpose IF data augmentation methods do not address. We propose \textbf{Reference-Grounded Data Curation}, a two-phase pipeline that extracts every supervised constraint from a reference translation that already satisfies it, ensuring feasibility by construction. Phase 1 applies Instruction-Following Difficulty (IFD) scoring to retain the hardest-but-learnable instances from an English-Thai parallel pool. Phase 2 extracts constraints from each reference target and keeps only generations satisfying every constraint, yielding the 1.97M-record Grounded dataset. We fine-tune open-weight bases on Grounded to produce \textbf{ChindaMT}, a Thai-English translation family at 4B, 2B, and 0.8B parameters. Under length-controlled pairwise judging, ChindaMT outperforms or matches every same-size baseline at every tier on both plain translation and under explicit rules, reaching up to a 68.4\% win rate against the strongest baseline. The recipe transfers cleanly across Qwen generations. We release model weights, the Grounded dataset, and evaluation suites.
\end{abstract}

\section{Introduction}

Large language models (LLMs) now translate competitively with dedicated MT systems \citep{xu2024paradigm, alves2024tower}, and translation requests increasingly carry prompt-level rules on terminology, formatting, register, and length that outputs must satisfy while staying accurate. Yet translation-specialized fine-tuning erodes prompt-following capability, while instruction-tuned LLMs without translation specialization follow such rules but underperform on quality. Open-weight Thai-English systems illustrate both failure modes. Thai-focused LLMs such as Typhoon \citep{pipatanakul2023typhoon, pipatanakul2024typhoon2} support instruction following (IF) but lack specialized translation quality, while multilingual machine translation (MT) specialists such as Hunyuan-MT \citep{tencent2025hunyuanmt} deliver translation quality without modeling prompt-level constraints. No open-weight Thai-English model currently closes this gap.

Two complementary lines of work build general IF training data. Data selection methods \citep{zhou2023lima, li-etal-2024-quantity, liu2024deita} identify a small high-quality SFT subset, while data augmentation methods \citep{wang-etal-2023-self-instruct, dong2025autoif, an-etal-2025-ultraif} synthesize constraint-bearing instructions by prompting an LLM. Both target general IF benchmarks such as IFEval \citep{zhou2023ifeval} and FollowBench \citep{jiang-etal-2024-followbench}, but neither has been adapted to translation.

We propose \textbf{Reference-Grounded Data Curation} (RGDC), a two-phase pipeline that grounds every supervised constraint in an existing reference translation. Phase 1 applies the Instruction-Following Difficulty (IFD) score \citep{li-etal-2024-quantity} to keep the hardest-but-learnable subset of an English-Thai parallel pool. Phase 2 extracts verifiable constraints from each retained reference, regenerates outputs, and keeps only candidates that pass every constraint under LLM-as-judge evaluation, producing the Grounded dataset.

We fine-tune four open-weight base models on Grounded to produce \textbf{ChindaMT}\footnote{Models: \href{https://huggingface.co/iapp}{\nolinkurl{huggingface.co/iapp/ChindaMT-{4B,2B,0.8B}}}. Grounded and the suites: \href{https://huggingface.co/iapp}{\nolinkurl{huggingface.co/datasets/iapp/ChindaMT-{Grounded,CoreEval,BroadEval}}}. Code: \href{https://github.com/iapp-technology/ChindaMT-RGDC}{\nolinkurl{github.com/iapp-technology/ChindaMT-RGDC}}.}, a Thai-English translation family at 4B, 2B, and 0.8B parameters. Three variants use Qwen3.5 bases and a fourth uses Qwen3-4B to demonstrate cross-generation transfer. Evaluation uses length-controlled pairwise win rate against same-size baselines on two suites we construct, with one primary and two open-weight cross-judges from independent model families. \textbf{CoreEval} covers five deployment-relevant domains spanning government, finance, tech, UI, and miscellaneous content, and \textbf{BroadEval} spans ten broader domains for cross-domain generalization. Each suite has a Plain split for translation only and a Constrained split with one to four rules.

On CoreEval, ChindaMT outperforms the size-matched MT specialists at all three sizes on both plain and constrained translation, and the same recipe transfers cleanly to Qwen3-4B as a cross-generation check. These wins extend to BroadEval and are corroborated by external metrics, two open-weight cross-judges from independent model families, and native-Thai human raters.

Our contributions are as follows:

\begin{itemize}
\item We propose \textbf{Reference-Grounded Data Curation}, which eliminates the compliance-quality conflict by extracting every constraint from a reference already satisfying it, so each rule-compliant target is a faithful translation.
\item We release \textbf{ChindaMT}, an open-weight Thai-English translation family at 4B, 2B, and 0.8B parameters, with the Grounded dataset and the \textbf{CoreEval} and \textbf{BroadEval} suites.
\item ChindaMT outperforms or matches every same-size baseline on plain and instruction-following translation, with external metrics, cross-judges, and humans agreeing and ablations isolating RGDC components.
\end{itemize}

\section{Related Work}

\subsection{Data Selection for Instruction Tuning}

LIMA \citep{zhou2023lima} showed that small high-quality SFT data can match much larger uncurated sets \citep{chen2024alpagasus, luo2023catastrophic}, motivating data-selection methods that retain high-signal instances. The IFD score \citep{li-etal-2024-quantity} selects samples where the instruction reduces but does not eliminate the loss on the output, yielding hard-but-learnable instances. Related methods include Deita \citep{liu2024deita}, Humpback \citep{li2024humpback}, and SuperFiltering \citep{li-etal-2024-superfiltering}.

We adapt IFD scoring to the translation setting because it defines difficulty with the loss of the base model being fine-tuned, so difficulty is measured for the model that will learn from the data, and needs no further scorer. SuperFiltering approximates this criterion with a weaker model to save cost. Deita scores complexity and quality, which have more room to separate samples in diverse open-domain pools than in our single-task pool.

\subsection[Instruction-Following Data Augmentation]{Instruction-Following\\ Data Augmentation}
\label{sec:if-data-aug}

Instruction-data augmentation methods synthesize new instruction-response pairs rather than selecting from an existing pool. Self-instruct \citep{wang-etal-2023-self-instruct} and WizardLM \citep{xu2023wizardlm} bootstrap diverse instructions from a seed set, while AutoIF \citep{dong2025autoif} and UltraIF \citep{an-etal-2025-ultraif} synthesize verifiable constraints from prompts and rejection-sample for compliance. RGDC inverts this synthesize-then-filter pattern, extracting constraints from existing reference translations so each is feasible by construction; Phase 2E filtering is then residual generation hygiene, not feasibility enforcement. A detailed comparison with AutoIF and UltraIF is in Appendix~\ref{app:related-comparison}, Table~\ref{tab:related-comparison}.

\subsection{Machine Translation with LLMs}

Recent work \citep{xu2024paradigm, xu2024cpo, alves2024tower, bawden-yvon-2023-investigating} shows that general-purpose LLMs can match dedicated neural MT systems. For Thai specifically, Typhoon \citep{pipatanakul2023typhoon, pipatanakul2024typhoon2}, SeaLLMs \citep{nguyen2023seallms, zhang2024seallmsv3}, and Hunyuan-MT \citep{tencent2025hunyuanmt} are representative recent efforts. Standard MT metrics like BLEU \citep{papineni-etal-2002-bleu}, chrF \citep{popovic-2015-chrf}, and COMET \citep{rei-etal-2020-comet} score translation quality but not rule compliance. Our work targets Thai-English IF-MT, where prior systems either support IF without specialized translation quality or deliver translation quality without modeling prompt-level constraints.

\subsection{LLM-as-Judge Evaluation}

LLM-as-judge protocols correlate well with human preferences \citep{zheng2023judging} but exhibit length bias \citep{saito2023verbosity} and position bias \citep{wang-etal-2024-large-language-models-fair}. AlpacaEval v2 \citep{dubois2024lcalpacaeval} addresses length bias via a generalized linear model fitted on each evaluation. We adopt this protocol with an open-weight 35B LLM judge for reproducibility, cross-validated by two further open-weight judges from independent model families (Section~\ref{sec:cross-judge}) and by native-Thai human raters.

\vspace*{0pt plus 80pt}

\section{Methodology}

\subsection{Pipeline Overview}

\begin{figure}[t]
  \centering
  \includegraphics[width=0.72\linewidth]{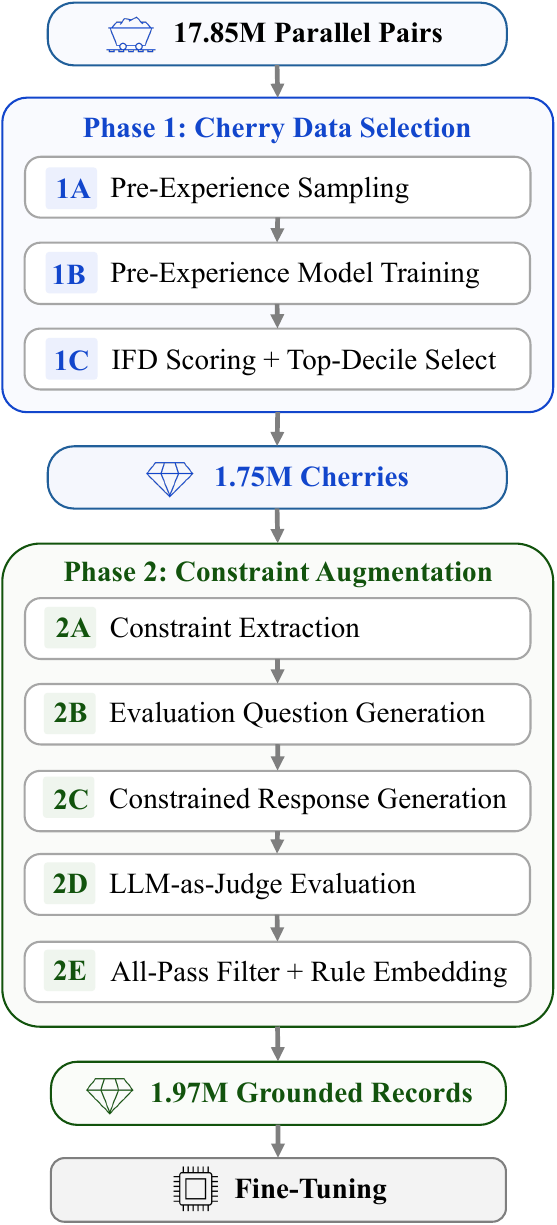}
  \caption{The two-phase RGDC pipeline. Phase 1 uses IFD scoring to cherry-select 1.75M instances from the 17.85M-record English-Thai parallel pool. Phase 2 extracts reference-grounded constraints and keeps only all-pass candidates, yielding the 1.97M-record Grounded dataset that trains the ChindaMT family.}
  \label{fig:rgdc-pipeline}
\end{figure}

\textbf{Reference-Grounded Data Curation} (RGDC) builds training data in two phases, shown in Figure~\ref{fig:rgdc-pipeline}. Phase 1 scores a large English-Thai parallel pool and keeps the hardest-but-learnable records. Phase 2 extracts verifiable constraints from each kept reference translation, so every constraint is feasible by construction, then regenerates the output under those constraints and keeps only generations that satisfy all of them. This yields the Grounded dataset, on which we fine-tune open-weight bases to produce the ChindaMT family.

\subsection{Phase 1: Cherry Data Selection}
\label{sec:phase1}

Our base model is Qwen3.5-4B \citep{qwen2025}. Following \citet{li-etal-2024-quantity}, we briefly fine-tune it on a representative subset, then score every pool record by IFD and keep the top decile.

\noindent\textbf{Pre-experience subset (1A--1B).} For each corpus-direction pair, the base model computes mean-pooled sentence embeddings and per-instance perplexity. We cluster the samples and draw a fixed quota from each cluster within a mid-perplexity band, avoiding trivially easy and noisy extremes. The base model is then briefly freeze-tuned on this subset, exposing it to the domain without overfitting individual instances.

\noindent\textbf{IFD scoring and selection (1C).} The pre-experienced model scores every record by IFD, its output-loss ratio with and without the instruction,
\begin{equation*}
\mathrm{IFD} = \frac{\ell(y \mid I, x)}{\ell(y)},
\end{equation*}
where $\ell$ is mean token-level cross-entropy, $y$ the target, $I$ the instruction, and $x$ the source. An IFD below 1 means the instruction eases prediction; at or above 1 it does not help, so we discard them. From the rest we keep the \textbf{top decile by IFD}, the hardest-but-learnable band, yielding $\sim$1.75M cherry samples (per-sub-step counts in Table~\ref{tab:record-counts}).

\subsection{Phase 2: Constraint Augmentation}
\label{sec:phase2}

As in AutoIF \citep{dong2025autoif} and UltraIF \citep{an-etal-2025-ultraif}, we use a larger auxiliary LLM for constraint extraction, regeneration, and judging. From the 1.75M cherry samples, we extract constraints from each reference target (2A), generate yes/no evaluation questions (2B), regenerate translations under those constraints (2C), judge each constraint (2D), and retain all-pass candidates (2E).

\noindent\textbf{Constraint extraction (2A).} For each cherry sample, the auxiliary LLM receives the source, the target translation, and a taxonomy of five constraint categories. It returns constraints the target already demonstrates in Table~\ref{tab:constraint-categories}, guaranteeing feasibility by construction. Full prompt in Appendix~\ref{app:prompts:2a}.

\begin{table}[h]
\centering
\footnotesize
\begin{tabularx}{\linewidth}{rX}
\toprule
\textbf{Category} & \textbf{Example} \\
\midrule
Content    & ``include the term `clean energy''' \\
Numerical  & ``use exactly one sentence'' \\
Stylistic  & ``use formal Thai without slang'' \\
Format     & ``return only the translation, no quotes'' \\
Linguistic & ``avoid English contractions'' \\
\bottomrule
\end{tabularx}
\caption{Example constraints for the five categories used by the Phase 2A extractor.}
\label{tab:constraint-categories}
\end{table}

\noindent\textbf{Evaluation question generation (2B).} All constraints from a sample are batched into a single call to the auxiliary LLM that returns one yes/no question per constraint. Full prompt in Appendix~\ref{app:prompts:2b}.

\noindent\textbf{Constrained response generation (2C).} Each cherry sample yields two candidate variants. The \textit{sampled} variant embeds 1--3 randomly selected constraints, and the \textit{all} variant embeds every extracted constraint. Each variant appends its constraints as a \texttt{Rules:} block to the source prompt, and the auxiliary LLM generates a new translation. System prompt in Appendix~\ref{app:prompts:2c}.

\noindent\textbf{LLM-as-judge evaluation (2D).} For each candidate variant, the auxiliary LLM answers the per-constraint yes/no questions from 2B, marking each satisfied or violated. These judgments drive the all-pass filter in 2E. Full prompt in Appendix~\ref{app:prompts:2d}.

\noindent\textbf{All-pass filtering (2E).} Only candidate variants where \textit{every} constraint receives YES are retained, yielding the \textbf{Grounded dataset} ($\sim$1.97M records; exact counts in Table~\ref{tab:record-counts}). The retained constraints are embedded into the training instruction as a \texttt{Rules:} block, producing clean (instruction, input, output) triples for supervised fine-tuning. A reference-free quality audit finds these regenerated targets at least as good as their source-corpus references (Appendix~\ref{app:record-example}).

\section{Experimental Setup}

\subsection{Source Parallel Corpora}
\label{sec:source-corpora}

Ten English-Thai parallel corpora (Appendix~\ref{app:corpora}, Table~\ref{tab:corpora}) supply both RGDC phases. After unification into Alpaca \texttt{(instruction, input, output)} format, the pool contains $\sim$17.85M records across both directions. Each source appears in both en-th and th-en for IFD scoring. Evaluation items drawn from this pool are held out of both phases from the outset (Section~\ref{sec:eval-data}).

\subsection{Grounded Training Dataset}
\label{sec:grounded}

The RGDC pipeline produces a single dataset we call \textbf{Grounded}, 1.97M records pairing each source and its \texttt{Rules:} block with the constraint-compliant translation. A worked record appears in Appendix~\ref{app:record-example}, with per-sub-step counts in Table~\ref{tab:record-counts}. Retained records carry 3.7 constraints on average, and $56.5\%$ of candidates survive to Grounded.

\begin{table}[h]
\centering
\small
\begin{tabular}{l r}
\toprule
\textbf{Sub-step} & \textbf{Records} \\
\midrule
\multicolumn{2}{l}{\textbf{Phase 1: Cherry selection}} \\
\quad 1C. Cherry-selected (top-decile)          & 1{,}751{,}962 \\
\multicolumn{2}{l}{\textbf{Phase 2: Constraint augmentation}} \\
\quad 2A. Constraint extractions                & 1{,}751{,}962 \\
\quad 2B. Evaluation questions                  & 1{,}747{,}959 \\
\quad 2C. Augmented prompts (2 variants)        & 3{,}495{,}704 \\
\quad 2E. All-pass filtered (\textbf{Grounded}) & \textbf{1{,}973{,}358} \\
\bottomrule
\end{tabular}
\caption{Record counts at each RGDC sub-step, starting from a 17.85M parallel pool. Sub-steps 1B and 2D do not change record counts.}
\label{tab:record-counts}
\end{table}

\subsection{Evaluation Datasets}
\label{sec:eval-data}

We construct two evaluation suites that share the Plain and Constrained formats but differ in data distribution, each with 400 samples per format, 200 en-th and 200 th-en. \textbf{CoreEval}, the primary suite, covers five deployment-relevant domains, government, finance, tech, UI, and miscellaneous, sampled from the parallel pool and length-stratified following \citet{pipatanakul2024typhoon2}. \textbf{BroadEval} is independently curated for generalization, spanning ten broader domains such as news, literary, and social text, with about $25\%$ paragraph-length inputs.

Both formats use English prompt scaffolding regardless of direction, matching the training format. Plain contains only the translation instruction and source, while Constrained adds a \texttt{Rules:} block of 1--4 constraints from the five categories in Table~\ref{tab:constraint-categories}. Full templates in Appendix~\ref{app:eval-prompts}.

\subsection{Models Compared}
\label{sec:models}

Our four ChindaMT variants share one fine-tuning recipe (Section~\ref{sec:hyperparams}) on Grounded, differing only in base. Three use Qwen3.5 at 4B, 2B, and 0.8B for the deployment scaling axis, plus Qwen3-4B for cross-generation transfer. We compare each against size-matched Thai-capable MT specialists, Typhoon-Translate-1.5 \citep{pipatanakul2024typhoon2}, Hunyuan-MT-1.5 \citep{tencent2025hunyuanmt}, GemmaX2-28 \citep{cui2025gemmax}, TranslateGemma \citep{finkelstein2026translategemma}, and MiLMMT-46 \citep{shang2026milmmt}, and against each variant's own base. Per-tier pairings are in Appendix~\ref{app:models-compared}, Table~\ref{tab:models}.

\subsection{Implementation Details}
\label{sec:hyperparams}

\noindent\textbf{Phase 1.} Sampling (1A) uses K-means with 100 clusters over base mean-pooled embeddings, drawing 10 per cluster within the ${25^{th}}$--${75^{th}}$ perplexity band ($40{,}683$ samples). Training (1B) freeze-tunes for 1 epoch with the top 4 transformer blocks unfrozen (FP32, AdamW lr $2 \times 10^{-5}$, inverse-square-root schedule, $1\%$ warmup, effective batch $64$).

\noindent\textbf{Phase 2.} The four LLM-using sub-steps share a single auxiliary LLM, Qwen3.5-35B-A3B-FP8 with thinking disabled, using the role-specific decoding parameters in Appendix~\ref{app:impl-hyperparams}, Table~\ref{tab:phase2-sampling}.

\noindent\textbf{Fine-tuning.} Full-parameter SFT for one epoch on Grounded, no preference optimization or RL. Core hyperparameters in Appendix~\ref{app:impl-hyperparams}, Table~\ref{tab:hyperparams}. The same recipe trains all four variants, with only the batch-size decomposition differing per tier ($4\times8$ for 4B and Qwen3-4B, $8\times4$ for 2B, $16\times2$ for 0.8B, each on two GPUs, preserving effective batch $64$). Runs in LLaMA-Factory \citep{zheng-etal-2024-llamafactory} on H100 under DeepSpeed ZeRO-2 with token packing and NEFTune noise $\alpha=2$.

\subsection{Evaluation Protocol}
\label{sec:eval-protocol}

\begin{table*}[t]
\centering
\small
\setlength{\tabcolsep}{4pt}
\begin{tabular}{cllrrrrrrrrr}
\toprule
& \multirow{2}{*}{\textbf{Model}} & \multirow{2}{*}{\textbf{Comparator}} & \multicolumn{3}{c}{\textbf{Plain (shared)}} & \multicolumn{3}{c}{\textbf{Plain (own)}} & \multicolumn{3}{c}{\textbf{Constrained}} \\
\cmidrule(lr){4-6} \cmidrule(lr){7-9} \cmidrule(lr){10-12}
& & & en$\rightarrow$th & th$\rightarrow$en & Mean & en$\rightarrow$th & th$\rightarrow$en & Mean & en$\rightarrow$th & th$\rightarrow$en & Mean \\
\midrule
\multirow{11}{*}{\rotatebox[origin=c]{90}{\textsc{CoreEval}}}
  & \multirow{5}{*}{\textbf{ChindaMT-4B}}   & MiLMMT-46-4B      & 88.8 & 83.7 & \textbf{86.3} & 62.8 & 42.1 & \textbf{53.3} & 89.9 & 88.7 & \textbf{89.5} \\
  &                                          & TranslateGemma-4B & 88.7 & 75.1 & \textbf{81.0} & 80.1 & 54.5 & \textbf{66.7} & 86.4 & 91.0 & \textbf{87.2} \\
  &                                          & Typhoon-Translate-4B    & 64.9 & 58.5 & \textbf{61.8} & 62.1 & 53.4 & \textbf{57.8} & 67.4 & 71.0 & \textbf{68.4} \\
  &                                          & HY-MT-1.5-7B      & 68.2 & 83.0 & \textbf{74.0} & 68.2 & 59.3 & \textbf{62.9} & 96.6 & 99.5 & \textbf{97.9} \\
  &                                          & GemmaX2-28-9B     & 94.7 & 92.8 & \textbf{93.6} & 66.2 & 45.0 & \textbf{56.6} & 94.4 & 93.5 & \textbf{94.1} \\
\cmidrule(l){3-12}
  & \multirow{3}{*}{\textbf{ChindaMT-2B}}   & MiLMMT-46-1B    & 88.5 & 88.7 & \textbf{88.4} & 62.9 & 47.5 & \textbf{54.6} & 93.0 & 94.3 & \textbf{93.7} \\
  &                                          & HY-MT-1.5-1.8B  & 73.9 & 79.5 & \textbf{75.2} & 78.3 & 70.9 & \textbf{72.6} & 82.7 & 80.7 & \textbf{78.6} \\
  &                                          & GemmaX2-28-2B   & 92.9 & 87.3 & \textbf{90.2} & 70.4 & 38.9 & \textbf{55.6} & 87.5 & 88.6 & \textbf{89.9} \\
\cmidrule(l){3-12}
  & \multirow{3}{*}{\textbf{ChindaMT-0.8B}} & MiLMMT-46-1B    & 85.9 & 81.3 & \textbf{83.4} & 55.6 & 47.9 & \textbf{50.9} & 93.5 & 87.3 & \textbf{90.3} \\
  &                                          & HY-MT-1.5-1.8B  & 62.9 & 69.7 & \textbf{64.6} & 67.7 & 63.5 & \textbf{63.4} & 74.1 & 68.8 & \textbf{67.8} \\
  &                                          & GemmaX2-28-2B   & 92.2 & 82.3 & \textbf{87.5} & 62.0 & 39.4 & \textbf{51.2} & 89.2 & 82.1 & \textbf{86.5} \\
\midrule
\multirow{11}{*}{\rotatebox[origin=c]{90}{\textsc{BroadEval}}}
  & \multirow{5}{*}{\textbf{ChindaMT-4B}}   & MiLMMT-46-4B      & ---  & ---   & ---           & 49.0 & 62.1 & \textbf{56.2} & ---  & ---   & ---           \\
  &                                          & TranslateGemma-4B & 80.4 & 83.7 & \textbf{81.8} & 72.5 & 62.5 & \textbf{67.0} & 74.6 & 80.3 & \textbf{76.1} \\
  &                                          & Typhoon-Translate-4B    & 60.0 & 56.6 & \textbf{58.7} & 53.3 & 49.8 & \textbf{53.1} & 63.4 & 63.1 & \textbf{63.2} \\
  &                                          & HY-MT-1.5-7B      & 51.3 & 87.8 & \textbf{70.1} & 57.5 & 68.5 & \textbf{62.0} & 96.7 & 99.7 & \textbf{98.2} \\
  &                                          & GemmaX2-28-9B     & ---  & ---  & ---           & 66.8 & 66.0 & \textbf{67.4} & ---  & ---  & ---           \\
\cmidrule(l){3-12}
  & \multirow{3}{*}{\textbf{ChindaMT-2B}}   & MiLMMT-46-1B    & 91.8 & 97.4 & \textbf{94.7} & 70.3 & 66.5 & \textbf{68.7} & ---   & ---  & ---           \\
  &                                          & HY-MT-1.5-1.8B  & 53.4 & 90.1 & \textbf{70.8} & 72.6 & 83.0 & \textbf{76.2} & 73.5 & 88.3 & \textbf{78.6} \\
  &                                          & GemmaX2-28-2B   & ---  & ---  & ---           & 67.3 & 63.9 & \textbf{66.3} & ---  & ---  & ---           \\
\cmidrule(l){3-12}
  & \multirow{3}{*}{\textbf{ChindaMT-0.8B}} & MiLMMT-46-1B    & 89.0 & 95.9 & \textbf{92.6} & 46.1 & 53.9 & \textbf{50.2} & ---  & ---  & ---           \\
  &                                          & HY-MT-1.5-1.8B  & 34.8 & 76.1 & \textbf{54.4} & 47.8 & 67.7 & \textbf{56.5} & 57.7 & 74.0 & \textbf{61.9} \\
  &                                          & GemmaX2-28-2B   & ---  & ---  & ---           & 44.8 & 51.2 & 49.4          & ---  & ---  & ---           \\
\bottomrule
\end{tabular}
\caption{Main pairwise LC\% on CoreEval and BroadEval. (shared) uses a uniform prompt scaffold, (own) lets each baseline use its own template. \textbf{Bold} marks Means above 50. Shared-prompt comparisons where the comparator returns almost no target-language output (GemmaX2 BroadEval, MM-46-4B BroadEval, and Constrained BroadEval for MM-46-1B at the 2B and 0.8B tiers) are omitted.}
\label{tab:main}
\end{table*}

\noindent\textbf{Judge.} We use a single judge, Qwen3.6-35B-A3B-FP8 with greedy decoding and thinking disabled, to score every pair in the AlpacaEval-v2 format under three criteria in priority order, accuracy and faithfulness, then instruction and format compliance, then fluency. Full prompt in Appendix~\ref{app:judge}.

\noindent\textbf{Metric.} From these preferences we report length-controlled win rate (LC\%) \citep{dubois2024lcalpacaeval}, where the AlpacaEval-v2 GLM corrects for length bias \citep{saito2023verbosity, wang-etal-2024-large-language-models-fair}, with standard errors defined in Appendix~\ref{app:metrics}. Raw win rates appear in Table~\ref{tab:main-direction}, external quality metrics in Section~\ref{sec:external-bench}.

\noindent\textbf{Judge-family bias.} The primary judge shares the Qwen family with ChindaMT and Typhoon-Translate, and LLM judges can mildly favor their own lineage \citep{zheng2023judging}, so Section~\ref{sec:cross-judge} re-judges every comparison with two open-weight judges from independent families.

\noindent\textbf{Split semantics.} Plain has only the translation instruction and source, so its win rate reflects translation quality and output format, with no explicit rule to follow (Table~\ref{tab:main}). Constrained adds a rules block, so its win rate additionally reflects explicit rule compliance, under the shared template only, since no baseline exposes a \texttt{Rules:} slot. Translation quality on its own is measured directly by the external metrics in Table~\ref{tab:external}.

\section{Results}
\label{sec:results}

\subsection{Main Pairwise Results}
\label{sec:main-results}

\looseness=-1 \noindent\textbf{CoreEval.} ChindaMT-4B outperforms every same-tier and larger baseline on Mean, on both plain and constrained translation, and whether baselines use the shared scaffold or their default templates. Against same-size Typhoon-Translate-4B it reaches LC\% $61.8$ on Plain (shared) and $68.4$ on Constrained, and Plain (own) LC\% stays above $50$ for every baseline, ruling out a prompt-format artifact. Constrained win rates over the larger HY-MT-1.5-7B approach the ceiling. ChindaMT-2B and ChindaMT-0.8B outperform or match every size-matched baseline, HY-MT-1.5-1.8B, GemmaX2-28-2B, and MiLMMT-46-1B. Per-direction breakdowns are in Appendix~\ref{app:per-direction}.

\noindent\textbf{BroadEval.} The CoreEval pattern holds out of domain. ChindaMT-4B and ChindaMT-2B lead every reportable size-matched comparison without a loss, ChindaMT-4B ahead of same-size Typhoon-Translate-4B with LC\% of $53.1$ to $63.2$ and of the larger HY-MT-1.5-7B by wider margins, while ChindaMT-0.8B leads its size-matched baselines except for a tie with GemmaX2-28-2B within $0.6$ of $50$. The instruction-following gap is the most durable out of domain, with Constrained LC over the larger HY-MT-1.5-7B reaching $98.2$.

\subsection{External-Benchmark Validation}
\label{sec:external-bench}

\begin{table*}[t]
\centering
\small
\setlength{\tabcolsep}{3pt}
\renewcommand{\arraystretch}{1.15}
\begin{tabular}{cclrrrrrrrrrr}
\toprule
& & \multirow{2}{*}{\textbf{Model}} & \multicolumn{2}{c}{\textbf{CK}$\uparrow$} & \multicolumn{2}{c}{\textbf{DA}$\uparrow$} & \multicolumn{2}{c}{\textbf{MQM}$\uparrow$} & \multirow{2}{*}{\shortstack{\textbf{Quality}\\\textbf{Mean}\,$\uparrow$}} & \multicolumn{2}{c}{\textbf{MtX}$\downarrow$} & \multirow{2}{*}{\shortstack{\textbf{MtX}\\\textbf{Mean}\,$\downarrow$}} \\
\cmidrule(lr){4-5} \cmidrule(lr){6-7} \cmidrule(lr){8-9} \cmidrule(lr){11-12}
& & & en$\rightarrow$th & th$\rightarrow$en & en$\rightarrow$th & th$\rightarrow$en & en$\rightarrow$th & th$\rightarrow$en & & en$\rightarrow$th & th$\rightarrow$en \\
\midrule
\multirow{11}{*}{\rotatebox[origin=c]{90}{\textsc{FLORES-200}}}
  & \multirow{6}{*}{\rotatebox[origin=c]{90}{\shortstack{\textsc{Larger}\\[-1pt]\scriptsize 4B--9B}}}
    & MiLMMT-46-4B      & \underline{83.88} & \textbf{84.98}    & 87.94             & \underline{91.90} & 97.01             & 98.23             & \textbf{90.7} & 2.14 & \textbf{1.97} & \underline{2.06} \\
  & & TranslateGemma-4B & 82.76             & 84.01             & 82.44             & 87.62             & 95.24             & 97.52             & 88.3 & 2.45 & 2.11 & 2.28 \\
  & & Typhoon-Translate-4B    & 83.62             & 83.97             & 87.52             & 87.79             & 97.23             & 97.74             & 89.6 & 2.14 & 2.26 & 2.20 \\
  & & HY-MT-1.5-7B      & \textbf{84.38}    & \underline{84.61} & \underline{88.71} & 89.48             & \textbf{97.63}    & 97.97             & \underline{90.5} & \textbf{1.98} & \underline{1.98} & \textbf{1.98} \\
  & & GemmaX2-28-9B     & 82.43             & 84.45             & 85.20             & \textbf{91.91}    & 96.94             & \underline{98.34} & 89.9 & 2.45 & 2.07 & 2.26 \\
  & & ChindaMT-4B (ours)& 83.80             & 84.55             & \textbf{89.32}    & 90.33             & \underline{97.53} & \textbf{98.39}    & \textbf{90.7} & \underline{2.13} & \underline{1.98} & \underline{2.06} \\
\cmidrule(l){2-13}
  & \multirow{5}{*}{\rotatebox[origin=c]{90}{\shortstack{\textsc{Smaller}\\[-1pt]\scriptsize 0.8B--2B}}}
    & MiLMMT-46-1B    & 82.56             & \textbf{84.41}    & 75.93             & 84.79             & 94.06             & 96.69             & 86.4 & 2.57 & 2.28 & \underline{2.43} \\
  & & HY-MT-1.5-1.8B  & \underline{83.03} & 83.65             & 79.39             & 82.64             & \underline{95.95} & 96.41             & 86.8 & \textbf{2.22} & \underline{2.25} & \textbf{2.24} \\
  & & GemmaX2-28-2B   & 81.80             & \underline{84.25} & \underline{79.78} & \textbf{88.06}    & 95.54             & \textbf{97.75}    & \underline{87.9} & 2.66 & \textbf{2.23} & 2.45 \\
  & & ChindaMT-0.8B (ours) & 82.51             & 83.41             & 78.71             & 79.54             & 95.27             & 95.66             & 85.9 & 2.50 & 2.55 & 2.53 \\
  & & ChindaMT-2B (ours)     & \textbf{83.40}    & 84.08             & \textbf{85.76}    & \underline{86.24} & \textbf{96.86}    & \underline{97.25} & \textbf{88.9} & \underline{2.25} & \textbf{2.23} & \textbf{2.24} \\
\midrule
\multirow{10}{*}{\rotatebox[origin=c]{90}{\textsc{WMT24++}}}
  & \multirow{6}{*}{\rotatebox[origin=c]{90}{\shortstack{\textsc{Larger}\\[-1pt]\scriptsize 4B--9B}}}
    & MiLMMT-46-4B      & 80.30             & \textbf{82.15}    & 84.39             & 89.21             & 94.74             & 96.68             & \underline{87.9} & \underline{3.14} & \underline{3.08} & \underline{3.11} \\
  & & TranslateGemma-4B & 78.55             & 80.51             & 77.21             & 84.31             & 93.14             & 95.92             & 84.9 & 3.37 & 3.28 & 3.33 \\
  & & Typhoon-Translate-4B    & 79.46             & 80.62             & 81.82             & 85.21             & 95.39             & 97.40             & 86.6 & 3.29 & 3.44 & 3.37 \\
  & & HY-MT-1.5-7B      & \textbf{81.15}    & \underline{81.90} & \textbf{87.19}    & \textbf{91.21}    & \textbf{96.67}    & \textbf{98.08}    & \textbf{89.4} & \textbf{2.62} & \textbf{2.80} & \textbf{2.71} \\
  & & GemmaX2-28-9B     & 77.64             & 81.10             & 75.72             & \underline{89.21} & 94.39             & \underline{97.65} & 86.0 & 3.95 & 3.32 & 3.64 \\
  & & ChindaMT-4B (ours)& \underline{80.23} & 81.21             & \underline{83.85} & 87.60             & \underline{96.63} & 97.36             & 87.8 & 3.21 & 3.19 & 3.20 \\
\cmidrule(l){2-13}
  & \multirow{5}{*}{\rotatebox[origin=c]{90}{\shortstack{\textsc{Smaller}\\[-1pt]\scriptsize 0.8B--2B}}}
    & MiLMMT-46-1B    & 77.84             & \textbf{81.17}    & 69.44             & 82.80             & 92.18             & 95.36             & 83.1 & 3.69 & 3.42 & 3.56 \\
  & & HY-MT-1.5-1.8B  & \textbf{79.58}    & \underline{80.93} & \underline{77.44} & \underline{83.94} & \underline{94.86} & \underline{96.16} & \textbf{85.5} & \textbf{2.93} & \textbf{3.08} & \textbf{3.01} \\
  & & GemmaX2-28-2B   & 77.06             & 80.81             & 70.77             & \textbf{84.99}    & 93.14             & \textbf{96.38}    & \underline{83.9} & 4.07 & 3.53 & 3.80 \\
  & & ChindaMT-0.8B (ours) & 77.80             & 79.58             & 69.65             & 74.28             & 92.05             & 94.14             & 81.2 & 3.74 & 3.71 & 3.73 \\
  & & ChindaMT-2B (ours)     & \underline{79.57} & 80.58             & \textbf{79.48}    & 82.25             & \textbf{95.02}    & 96.12             & \textbf{85.5} & \underline{3.30} & \underline{3.39} & \underline{3.35} \\
\bottomrule
\end{tabular}
\caption{External-metric translation quality on FLORES-200 (Wikipedia) and WMT24++ en-th (news), with CometKiwi (CK), GEMBA-DA (DA), and GEMBA-MQM (MQM) as $0$--$100$ quality scores ($\uparrow$); \textbf{Mean}$\uparrow$ is their unweighted six-value average. MetricX-24 (MtX, $\downarrow$) is the native error score on a $0$--$25$ scale (lower better), kept separate with its own \textbf{Mean}$\downarrow$ over the two directions. Rows ordered by size within each tier, ChindaMT last. \textbf{Bold} marks the best per column within a tier, underline the second, ties bolded jointly.}
\label{tab:external}
\end{table*}

We test whether the LC gains cost raw translation quality on FLORES-200 devtest (Wikipedia) and WMT24++ en-th (news), reporting four reference-free and reference-based metrics that correlate with human judgment, CometKiwi \citep{rei-etal-2022-cometkiwi}, GEMBA-DA \citep{kocmi-federmann-2023-large}, GEMBA-MQM \citep{kocmi-federmann-2023-gemba}, and MetricX-24 \citep{juraska-etal-2024-metricx}. GEMBA-DA and GEMBA-MQM reuse our Qwen3.6-35B judge, with definitions and prompts in Appendices~\ref{app:metrics} and~\ref{app:metric-prompts}. Each model runs under its best-case prompt, ChindaMT with the Grounded format and each baseline with its own model-card template.

ChindaMT-4B ties the top of FLORES Larger Mean (90.7, with MiLMMT-46-4B) and outperforms same-size Typhoon-Translate-4B on most comparisons across both benchmarks, while ChindaMT-2B leads its tier on FLORES (88.9) and ties for the WMT24++ lead (85.5). On WMT24++ Larger, ChindaMT-4B (87.8) trails HY-MT-1.5-7B (89.4) and sits level with MiLMMT-46-4B (87.9), the 1.6-point gap falling on the news-domain th-en direction where the larger 7B specialist holds an edge. Pairwise judging (Table~\ref{tab:main}) shows ChindaMT-2B leads size-matched comparisons, so external-metric strength does not cost instruction following. Full numbers in Appendix~\ref{app:per-direction}, Table~\ref{tab:main-direction}.

\subsection{Cross-Judge and Human Validation}
\label{sec:cross-judge}

We check that LC rankings are not a primary-judge artifact, via cross-judge re-evaluation across independent families and blinded native-Thai rating.

\noindent\textbf{Cross-judge.} We re-judge every comparison in Table~\ref{tab:main} with two additional open-weight judges from distinct families, GPT-OSS-120B \citep{openai2025gptoss} and Llama-3.3-70B-Instruct \citep{grattafiori2024llama3}, neither sharing lineage with any baseline. Constrained agreement is decisive, with ChindaMT WR reaching the mid-90s against the strongest MT specialists on both suites and judge spread under three points. Plain is judge-sensitive on MM-46 and GemmaX2 (WR\% in $[45, 74]$), consistent with their parity on external metrics (Table~\ref{tab:external}). All 16 Constrained and 18 Plain (shared) comparisons agree on direction, and the three direction-disagreeing comparisons of 56 are near-parity Plain (own) within five points of 50\%.

\begin{table}[t]
\centering
\small
\setlength{\tabcolsep}{6pt}
\begin{tabular}{llrr}
\toprule
\textbf{Model} & \textbf{Comparator} & \textbf{Plain} & \textbf{Constr.} \\
\midrule
\multirow{5}{*}{\textbf{ChindaMT-4B}}
  & MM-46-4B   & 51.0 & 89.6 \\
  & TG-4B      & 63.3 & 85.1 \\
  & Typhoon-4B & 56.1 & 65.5 \\
  & HY-MT-7B   & 57.2 & 95.3 \\
  & GemmaX2-9B & 56.4 & 93.7 \\
\cmidrule(l){2-4}
\multirow{3}{*}{\textbf{ChindaMT-2B}}
  & MM-46-1B   & 54.1 & 91.3 \\
  & HY-MT-1.8B & 62.1 & 73.4 \\
  & GemmaX2-2B & 56.0 & 91.9 \\
\cmidrule(l){2-4}
\multirow{3}{*}{\textbf{ChindaMT-0.8B}}
  & MM-46-1B   & 51.7 & 88.2 \\
  & HY-MT-1.8B & 57.0 & 65.1 \\
  & GemmaX2-2B & 53.0 & 90.1 \\
\bottomrule
\end{tabular}
\caption{Cross-judge ChindaMT WR\% on CoreEval, averaged across the three open-weight judges. Plain uses each comparator's own template, and values above 50 favor ChindaMT. Comparators are abbreviated, with full names in Table~\ref{tab:models}. BroadEval, Plain (shared), and per-judge numbers in Appendix~\ref{app:cross-judge-full}.}
\label{tab:cross-judge}
\end{table}

\noindent\textbf{Human validation.} Three native Thai raters scored blinded Plain and Constrained pairs from the ChindaMT-4B against Typhoon-Translate-1.5-4B comparison, Plain with the baseline template and Constrained with the shared template, full protocol in Appendix~\ref{app:human}. ChindaMT-4B is preferred on both splits, with stronger intensity on Constrained ($+0.68$) than Plain ($+0.39$). Inter-rater $\kappa$ averages $0.73$ (substantial) and judge-vs-human-majority $\kappa = 0.41$ (moderate), supporting the LLM judge as a population-level proxy. Agreement is stronger on Constrained ($0.62$) than on Plain ($0.18$), where the own-template condition leaves the two systems close (Table~\ref{tab:main}) and little signal for either judge or rater to separate.

\begin{table}[htbp]
\centering
\small
\setlength{\tabcolsep}{4pt}
\begin{tabular}{lrrr}
\toprule
\textbf{Metric} & \textbf{Plain} & \textbf{Constrained} & \textbf{Overall} \\
\midrule
ChindaMT win rate          & 0.64    & 0.69    & 0.67    \\
Pref.\ intensity           & $+0.39$ & $+0.68$ & $+0.53$ \\
Inter-rater $\kappa$       & 0.74    & 0.70    & 0.73    \\
Human-vs-judge $\kappa$    & 0.18    & 0.62    & 0.41    \\
\bottomrule
\end{tabular}
\caption{Human validation by three native Thai raters on 50 Plain and 50 Constrained items from the ChindaMT-4B against Typhoon-Translate-1.5-4B comparison, Plain with the baseline template and Constrained with the shared template. Win rate averages ChindaMT preference across raters (ties as half-wins), and intensity is the mean $-2$ to $2$ rating. Inter-rater $\kappa$ averages three pairs, and human-vs-judge $\kappa$ compares against the human majority.}
\label{tab:human}
\end{table}

All four channels, the primary LC\% (Table~\ref{tab:main}), external metrics (Table~\ref{tab:external}), three-judge cross-judge (Table~\ref{tab:cross-judge}), and human evaluation (Table~\ref{tab:human}), agree on direction, so the headline does not rest on any single validation channel.

\subsection{Method Lift and Parameter Scaling}
\label{sec:own-base}

We measure two scaling axes, own-base lift and within-Qwen3.5 parameter scaling at fixed recipe (Appendix~\ref{app:scaling}, Table~\ref{tab:own-base}). All four Qwen fine-tunes outperform their bases on CoreEval, with Plain/Constrained LC\% of $55.7/62.2$ (Qwen3.5-4B), $71.9/69.5$ (Qwen3-4B), $77.1/74.2$ (Qwen3.5-2B), and $81.5/83.5$ (Qwen3.5-0.8B). Lift grows as parameters decrease, and Qwen3-4B gains more than Qwen3.5-4B from the same supervision, reflecting more headroom in smaller and older bases \citep{zhou2023lima, chen2024alpagasus}.

Within-family scaling is asymmetric. Plain lifts compress as size gaps narrow ($+12.6$ for 4B vs 2B, $+6.9$ for 2B vs 0.8B), while Constrained lifts stay roughly constant per halving ($+14.2$ and $+14.4$). IF capability therefore scales sub-linearly and is largely preserved at 2B, making ChindaMT-2B a useful intermediate deployment choice.

\noindent\textbf{Cross-family transfer.} Fine-tuned on Grounded with only the base and chat template changed, the Gemma-3-4B fine-tune (FT) outperforms the Gemma-3-4B base with Plain/Constrained LC\% of $99.1/66.6$ (Table~\ref{tab:own-base}). All three judges agree on direction. The Plain margin is near ceiling because the base embeds its translation in English commentary that the rubric penalizes. The Constrained margin additionally reflects rule compliance. Since Phase 1 selection used the Qwen3.5-4B scorer, Grounded transfers across model families without base-specific selection.

\subsection{Component Ablations}
\label{sec:ablations}

Each Qwen3.5-4B ablation drops one RGDC component from the full recipe and shares all other hyperparameters with ChindaMT-4B (Table~\ref{tab:ablation-lc}). \textit{No-cherry} replaces IFD selection with uniform random sampling from the 17.85M pool, \textit{no-all-pass} drops the Phase 2E filter, and \textit{no-rules} strips the \texttt{Rules:} block from training input.

\begin{table}[t]
\centering
\small
\setlength{\tabcolsep}{5pt}
\begin{tabular}{lcccrr}
\toprule
& \multicolumn{3}{c}{\textbf{Components}} & \multicolumn{2}{c}{\textbf{Full recipe lift}} \\
\cmidrule(lr){2-4} \cmidrule(lr){5-6}
\textbf{Variant} & Cherry & Filter & Rules & Plain & Constr. \\
\midrule
Full         & $\checkmark$ & $\checkmark$ & $\checkmark$ & ---      & ---      \\
no-cherry    & $\times$     & $\checkmark$ & $\checkmark$ & $+8.24$  & $+0.69$  \\
no-all-pass  & $\checkmark$ & $\times$     & $\checkmark$ & $+1.09$  & $+1.06$  \\
no-rules     & $\checkmark$ & $\checkmark$ & $\times$     & $+12.63$ & $+9.53$  \\
\bottomrule
\end{tabular}
\caption{RGDC component ablations at 4B on CoreEval. Full recipe lift is the LC\% margin over each ablation in pairwise judging, above the $50$ tie. SE $[2.1, 2.5]$.}
\label{tab:ablation-lc}
\end{table}

\noindent\textbf{Cherry selection contributes 8 LC on Plain only.} The full recipe lifts $+8.24$ over no-cherry on Plain (outside SE) and $+0.69$ on Constrained (within SE). External metrics on out-of-domain FLORES and WMT24++ stay around $\pm 0.3$ on dataset Means (Appendix~\ref{app:ablation-external}), so Phase 1 aligns outputs with the CoreEval distribution rather than raising translation quality uniformly. Constraint compliance comes from Phase 2 supervision.

\noindent\textbf{All-pass filter contributes a small positive lift.} The full recipe gains $+1.09/+1.06$ over no-all-pass, modest relative to SE but consistent in sign on both axes, with external-metric deltas small on both benchmarks (Appendix~\ref{app:ablation-external}). The filter is data hygiene that complements reference grounding.

\noindent\textbf{Rule-block training contributes 10--13 LC.} The full recipe lifts $+12.63/+9.53$ over no-rules, above SE on every comparison, and no-rules loses all six quality-metric comparisons with Mean down $0.7$--$0.9$ (Appendix~\ref{app:ablation-external}, Table~\ref{tab:ablation-ext}). No-rules hurts Plain at least as much as Constrained, so rule-conditioned supervision aligns the broader output distribution with high-quality reference style rather than a narrow IF mechanism.

\section{Analysis}

\subsection{Translation--IF Trade-off}
\label{sec:tradeoff}

Translation-specialized models forfeit instruction-following as they gain translation quality \citep{xu2024paradigm, alves2024tower}, and the pattern recurs across our baselines. Tables~\ref{tab:external} and~\ref{tab:cross-judge} show MM-46-4B, TranslateGemma-4B, and GemmaX2 at parity with ChindaMT on plain translation, yet they lose every Constrained comparison under all three judges. Typhoon-Translate-4B occupies the opposite corner, with weaker plain translation and a smaller Constrained gap.

RGDC avoids both failure modes because rule and translation supervision share one source of truth, the reference translations that cherry selection retains. The Phase 2 ablation (Section~\ref{sec:ablations}) supports the mechanism, since removing the Rules block reduces Plain LC more than Constrained LC, pointing to broad output-distribution alignment rather than a narrow instruction-following circuit. Plain-translation parity with the strongest size-matched MT specialists is the central evidence that the instruction-following gains do not cost translation quality.

On external metrics ChindaMT reaches parity rather than dominance, the one residual specialist edge being in-domain news, where the larger 7B HY-MT model leads on WMT24++ th-en (Table~\ref{tab:external}). The deployment-relevant gains, rule compliance and cross-domain robustness at a fixed size, are where ChindaMT leads.

\subsection{Qualitative Case Analysis}
\label{sec:qualitative}

Consider a Thai-to-English translation under the rule \textit{``Return only the translated text''}. ChindaMT-4B and its Qwen3.5-4B base both follow the rule; Typhoon-Translate echoes the rule block and an \texttt{EN:} prefix before its translation, a direct rule violation. BLEU, chrF, and neural QE metrics would partially accept the Typhoon output because the trailing translation is correct, while LC\% catches the format violation through the hard penalty in Appendix~\ref{app:judge}. Full transcript in Appendix~\ref{app:qualitative-case}.

This case instantiates the two failure modes that motivate the paper. Typhoon-Translate, a translation specialist, renders the sentence accurately but leaks the rule block and an \texttt{EN:} prefix, quality without compliance. The instruction-tuned Qwen3.5-4B base obeys the rule yet drops the source word \textit{web}, compliance with a faithfulness loss. ChindaMT-4B alone does both, returning only the translation while preserving the content its base omits, resolving the trade-off in Section~\ref{sec:tradeoff}.

\section{Conclusion}

We introduced \textbf{ChindaMT}, an open-weight in\-struc\-tion-fol\-low\-ing Thai-English trans\-la\-tion family at 4B, 2B, and 0.8B parameters, and \textbf{RGDC}, the two-phase pipeline that builds its training data. RGDC extracts every rule from a reference translation that already satisfies it, so feasibility is guaranteed by construction. Under length-controlled pairwise evaluation, ChindaMT outperforms or matches size-matched translation specialists at every tier on CoreEval and BroadEval, without sacrificing raw translation quality on FLORES-200 or WMT24++, a result that two open-weight cross-judges and native Thai raters corroborate. The recipe transfers across four bases and two Qwen generations. We release model weights, the Grounded dataset, evaluation suites, and code under open licenses.

\section{Limitations}

\noindent\textbf{Reference-bounded constraint coverage.} RGDC extracts only rules a reference already satisfies, so the constraints it can supervise are bounded by the reference distribution. Demands that no reference exemplifies, or that resist a yes/no check, fall outside the pipeline, trading breadth for feasibility. Term, length, register, and output-format instructions are covered. A mandated glossary and placeholder tokens fall outside by design, since reference grounding supervises only what a reference demonstrates. Widening coverage by grounding in the resources of a domain, or with synthetic constraint generation, is left to future work.

\noindent\textbf{Single primary judge.} Primary LC and GEMBA rely on one open-weight judge (Qwen3.6-35B), so absolute LC values may reflect its calibration. We mitigate this with external metrics (Table~\ref{tab:external}), AlpacaEval-v2 length and position controls, a three-family cross-judge (Table~\ref{tab:cross-judge}), and human raters (Table~\ref{tab:human}). Direction agreement holds across all three judge families on every Constrained and Plain (shared) comparison.

\noindent\textbf{Language-pair and domain coverage.} All experiments cover English-Thai in both directions only. Extension to other low-resource pairs, especially Southeast Asian languages, and to broader registers is left to future work.

\noindent\textbf{Prompt-format asymmetry.} The Grounded training format matches the Typhoon-Translate template but differs from HY-MT and GemmaX2. On Plain we evaluate baselines under both shared and own templates. Constrained uses the shared format only, since no baseline exposes a \texttt{Rules:} slot, omitting some GemmaX2 comparisons (Table~\ref{tab:main}).

\noindent\textbf{Base-architecture coverage.} All four ChindaMT bases are Qwen. The recipe transfers across Qwen3.5 and Qwen3 generations, and the curated data lifts Gemma-3-4B over its base (Table~\ref{tab:own-base}), though with Qwen-derived Phase 1 selection. Rerunning selection with the target base and transfer to other architectures remain untested.

\section{Ethical Considerations}

\noindent\textbf{Data.} RGDC uses ten public English-Thai corpora (Appendix~\ref{app:corpora}). Because several carry ShareAlike terms, the released Grounded dataset and evaluation suites adopt CC-BY-SA 4.0. The suites ship source text and model outputs but no reference translations or personally identifying information, consistent with the source licenses, and one BroadEval item retains the CC-BY-NC-4.0 terms of its source, marked in the released file.

\noindent\textbf{Human evaluation.} The study (Appendix~\ref{app:human}) used three native Thai speakers who volunteered without compensation and gave informed consent after being told the purpose and their right to withdraw. They rated anonymized pairs in a minimal-risk task recording only preferences, with no personal data collected. The study required no formal ethics-board review at our institution.

\noindent\textbf{Intended use and risks.} Like any MT system, ChindaMT can produce fluent but incorrect output or reflect training-data bias, so its outputs warrant human review in high-stakes use.

\noindent\textbf{Release and compute.} Model weights and code are released under Apache-2.0, the weights inheriting the Qwen base-model license, and the Grounded dataset and evaluation suites under CC-BY-SA 4.0, for reproducibility. The small 0.8B--4B models keep compute modest and accessible to the low-resource Thai-English community.

\section*{Acknowledgements}
This research is financially supported under the Thammasat University's research fund, Center of Excellence in Intelligent Informatics, Speech and Language Technology and Service Innovation (CILS), and Intelligent Informatics and Service Innovation (IISI) Research Center, the Thailand Research Fund under grant number RTA6080013, as well as National Research Council of Thailand (NRCT), Hub of Talents in AI and Innovation Next (AI-Next), N35E690166.
We thank the anonymous reviewers and the area chair for their constructive comments, SiamAI for computational resources, OpenThai Lab for its support, and the three native Thai speakers who volunteered as raters. An AI assistant was used for language editing, drafting support, and help with LaTeX and code. The authors reviewed and verified all content and take full responsibility for it.

\bibliography{custom_camera}

\appendix

\section{Detailed Comparison with Prior IF-Data-Augmentation Methods}
\label{app:related-comparison}

Expanding Section~\ref{sec:if-data-aug}, Table~\ref{tab:related-comparison} compares RGDC against the two closest IF-data-augmentation methods, AutoIF \citep{dong2025autoif} and UltraIF \citep{an-etal-2025-ultraif}, across seven axes.

The axes isolate where RGDC departs from synthesize-then-filter methods. Extracting constraints from a reference that already satisfies them makes every target feasible by construction, whereas AutoIF and UltraIF synthesize from prompts and discard the unsatisfiable share. This admits an all-pass filter stricter than thresholding or rejection sampling, and grounding in a large parallel pool scales RGDC to 1.97M translation-specialized records, an order of magnitude beyond their general-domain corpora.

\section{RGDC Pipeline Prompts}
\label{app:prompts}

All four prompts run on vLLM with \texttt{enable\_thinking=false} and per-role sampling from Table~\ref{tab:phase2-sampling}, with auxiliary LLM \texttt{Qwen3.5-35B-A3B-FP8} (Section~\ref{sec:hyperparams}). The four prompts chain into one Phase 2 record, with 2A extracting constraints, 2B turning each into a verification question, 2C regenerating under them, and 2D scoring compliance.

\subsection{Constraint Extraction (2A)}
\label{app:prompts:2a}

Extracts verifiable constraints across five categories from a Simplified Query and its Answer (the existing reference translation). The five categories match the evaluation-time taxonomy (Table~\ref{tab:constraint-categories}), and requiring each constraint to be verifiable and absent from the query keeps it a checkable signal rather than a restatement of the task. The full template follows, with its five in-context examples.

\begin{lstlisting}
You are an expert in generating instruction constraints for a given simplified query.

Definition of Constraint: The smallest unit of restriction or requirement that can be added to an instruction to make the task more specific or challenging.

Your Task: Given a Simplified Query and its Answer, generate NEW constraints that could be added to the query. The generated constraints should be relevant to the query and the answer, and should make the task more specific without changing its fundamental goal.

Simplified Query: {QUERY}
Answer: {ANSWER}

Follow these steps:
1. Understand the basic goal of the simplified query.
2. Analyze the answer to understand what aspects can be constrained.
3. Generate NEW constraints that are:
   - Relevant to the query and answer
   - Verifiable in the response
   - Not already explicitly stated in the query
4. Categorize constraints into the following types:

Constraint Types:
- Content Constraints:
    - Specific Terms or Symbols: Mandatory use of certain terms or symbols (e.g., must include the word 'beautiful').
    - Required Elements or Concepts: Specific elements or concepts that must be included (e.g., must mention the Great Wall).
    - Thematic Directives: Instructions about thematic content, perspective, or focus (e.g., focus on environmental impact).

- Numerical Constraints:
    - Quantities related to the content: number of points, sentences, paragraphs, word count, or examples (e.g., exactly three sentences, at least 5 examples).

- Stylistic Constraints:
    - Tone and style requirements (e.g., formal, informal, conversational, humorous).
    - Specific language or terminology preferences (e.g., encyclopedic style, avoid jargon).

- Format Constraints:
    - Structure or format requirements (e.g., list, JSON, bullet points, table).
    - Programming language specifications (e.g., Java, Python).
    - Presentation styles (e.g., markdown, code block format).

- Linguistic Constraints:
    - Language specifications (e.g., in English, in Spanish).
    - Sentence structure requirements (e.g., use only simple sentences, imperative form).
    - Word-level requirements (e.g., lowercase, single-rhyme, alliteration).
 
Response Format:
- Generate constraints that are appropriate for the given query.
- Ensure generated constraints do not overlap with what is already in the query.
- Only include constraint types that are relevant.
- Present each constraint as a dictionary with a 'constraint' key.

```json
{
    "Content Constraints": [
        {"constraint": "..."},
        {"constraint": "..."}
    ],
    "Numerical Constraints": [...],
    "Stylistic Constraints": [...],
    "Format Constraints": [...],
    "Linguistic Constraints": [...]
}
```

Examples:

---
Example 1:
Query: "How do I check if a user pressed the cancel button on a prompt in JS?"
Answer: "You can check if a user pressed cancel by comparing the prompt result to null. When the user clicks cancel, the prompt() function returns null instead of a string. So you simply use: if (result === null) { /* user cancelled */ }"

Response:
```json
{
    "Numerical Constraints": [
        {"constraint": "Answer with exactly three sentences."}
    ]
}
```
Explanation: The query asks for a JavaScript explanation. A numerical constraint "exactly three sentences" is generated to add structure to the response. This constraint is verifiable by counting the sentences in the answer (which has exactly 3 sentences).

---
Example 2:
Query: "Explain quantum computing"
Answer: "Quantum computing is like having a super-powered calculator that can try many answers at once instead of one at a time. Regular computers use bits (0s and 1s), but quantum computers use 'qubits' that can be both 0 and 1 at the same time. This lets them solve certain complex problems much faster than regular computers."

Response:
```json
{
    "Content Constraints": [
        {"constraint": "Explain in simple terms using analogies."}
    ]
}
```
Explanation: The query asks for an explanation of quantum computing. A content constraint "explain in simple terms using analogies" is generated because the answer uses accessible language and analogies (like "super-powered calculator"). This makes the explanation requirement more specific without changing the core goal.

---
Example 3:
Query: "I want to replace string like 'This is a new {object} at {place}' with a Map, {object: 'student', place: 'point 3, 4'}, and get a result 'This is a new student at point 3, 4'. How can I do this?"
Answer: "public String replaceWithMap(String template, Map<String, String> values) {\n    String result = template;\n    for (Map.Entry<String, String> entry : values.entrySet()) {\n        result = result.replace('{{' + entry.getKey() + '}}', entry.getValue());\n    }\n    return result;\n}"

Response:
```json
{
    "Format Constraints": [
        {"constraint": "Answer using Java programming language."}
    ]
}
```
Explanation: The query asks how to perform string replacement. The answer is provided in Java code. A format constraint "Answer using Java" is generated to specify the programming language requirement, making the query more precise about the expected output format.

---
Example 4:
Query: "Write a program which accepts a command line parameter as a question and sends it to a server via HTTP GET method."
Answer: "import sys\nimport requests\n\nif len(sys.argv) < 2:\n    print('Usage: python script.py <question>')\n    sys.exit(1)\n\nquestion = sys.argv[1]\nresponse = requests.get('http://example.com/api', params={'question': question})\nprint(response.text)"

Response:
```json
{
    "Format Constraints": [
        {"constraint": "Write as a Python program."}
    ]
}
```
Explanation: The query asks for a program without specifying the language. Since the answer is in Python, a format constraint "Write as a Python program" is generated to make the language requirement explicit. This ensures the response will be in Python.

---
Example 5:
Query: "Recommend books"
Answer: "• 'Dream of the Red Chamber' by Cao Xueqin - A classic tale of noble families\n• 'Journey to the West' by Wu Cheng'en - Adventures of a monk and his companions\n• 'Romance of the Three Kingdoms' by Luo Guanzhong - Epic historical warfare\n• 'Water Margin' by Shi Nai'an - Tales of outlaws and heroes\n• 'The Scholar' by Wu Jingzi - Satire of the imperial examination system\n• 'Strange Tales from a Chinese Studio' by Pu Songling - Collection of supernatural stories\n• 'Border Town' by Shen Congwen - A love story set in rural China\n• 'Family' by Ba Jin - Story of generational conflict\n• 'Rickshaw Boy' by Lao She - Tale of a rickshaw puller's struggles\n• 'To Live' by Yu Hua - A man's journey through tumultuous times"

Response:
```json
{
    "Stylistic Constraints": [
        {"constraint": "Use Shakespeare's tone."}
    ],
    "Format Constraints": [
        {"constraint": "Use bullet points in the answer."}
    ],
    "Numerical Constraints": [
        {"constraint": "Recommend exactly ten books."}
    ],
    "Content Constraints": [
        {"constraint": "Focus on Chinese literature."}
    ]
}
```
Explanation: The query simply asks to "recommend books" which is very broad. Multiple constraints are generated based on the answer: (1) Stylistic constraint for "Shakespeare's tone" adds a creative challenge; (2) Format constraint for "bullet points" matches the answer's structure; (3) Numerical constraint for "ten books" matches the count in the answer; (4) Content constraint for "Chinese literature" narrows the scope to match the answer's focus.

---

Please only provide the response in JSON format.
\end{lstlisting}

\subsection{Evaluation Question Generation (2B)}
\label{app:prompts:2b}

\looseness=-1 Converts each constraint from Appendix~\ref{app:prompts:2a} into a yes/no verification question, batched into one LLM call per record. Single yes/no questions make compliance objectively checkable in Phase 2D, and the empty-string fallback drops descriptive or embedded constraints so the filter acts only on genuine requirements.

\begin{lstlisting}
You are an expert in crafting questions to evaluate whether a response to a query adheres to specific constraints.

For each constraint below, design a question that human evaluators can use to assess if the response meets that constraint. Each question should focus solely on its given constraint.

If a constraint is meaningless or is part of the content itself (e.g., descriptions, scenarios, examples), respond with an empty string for that constraint.

Example:
Query: Recommend books.
Constraints:
1. Use bullet points in your answer.
2. Recommend exactly ten books.

Response:
{
    "1": "Does the response use bullet points?",
    "2": "Does the response recommend exactly ten books?"
}

Now generate evaluation questions for:
Query: {query}

Constraints:
{constraints_list}

Respond only in JSON format mapping constraint numbers to questions:
{
    "1": "question for constraint 1",
    "2": "question for constraint 2"
}
\end{lstlisting}

\begin{table*}[t]
\centering
\footnotesize
\begin{tabular}{@{}lp{3.5cm}p{3.7cm}p{5.0cm}@{}}
\toprule
\textbf{Aspect} & \textbf{AutoIF} & \textbf{UltraIF} & \textbf{RGDC (ours)} \\
\midrule
Source data         & 36 hand-written seeds              & ShareGPT prompts                       & Cherry-selected translation Q\&A \\
\addlinespace
Constraint origin   & Self-instruct from seeds & Decomposed from prompts & Extracted from references \\
\addlinespace
Verification        & Python execution                   & LLM-as-judge                           & LLM-as-judge with explanations \\
\addlinespace
Filter strictness   & Score threshold                    & Rejection sampling                     & Every constraint must pass \\
\addlinespace
Scale               & 10--25k                            & 175k                                   & 1.97M \\
\addlinespace
Aux.\ model         & No                                 & Yes (UltraComposer)                    & Yes (IFD scorer + aux LLM) \\
\addlinespace
Domain              & General                            & General                                & Translation \\
\bottomrule
\end{tabular}
\caption{Comparison of RGDC with AutoIF and UltraIF across seven axes.}
\label{tab:related-comparison}
\end{table*}

\subsection{Constrained Response Generation (2C)}
\label{app:prompts:2c}

Each cherry record produces two variants, \textit{sampled} (1--3 constraints) and \textit{all} (every extracted constraint). Each is formatted with a \texttt{Rules:} block appended to the original instruction and sent to the generator with the system prompt below. The two variants expose the model to varying rule counts, and the system prompt suppresses preamble so each target stays a clean translation.

\needspace{4\baselineskip}
\noindent\textbf{System prompt.}

\begin{lstlisting}
You are an expert tasked with answering the given query. Please provide a clear and concise response directly, without introductory phrases such as 'What a great question,' 'Here is the answer,' or similar expressions. Focus solely on addressing the query while strictly following its inside constraints.
\end{lstlisting}

\noindent\textbf{User message template.}

\begin{lstlisting}
[Original Instruction]

Rules:
1. [Constraint 1]
2. [Constraint 2]

[Original Input]
\end{lstlisting}

\subsection{Compliance Evaluation (2D)}
\label{app:prompts:2d}

\looseness=-1 The judge evaluates every yes/no question from Appendix~\ref{app:prompts:2b} against each candidate from Appendix~\ref{app:prompts:2c}, returning \texttt{YES}/\texttt{NO} with explanation. Records pass the Phase 2E filter only if every question receives \texttt{YES}. Requiring a justification before each verdict reduces label noise, which matters because a single \texttt{NO} drops the record under the all-pass filter.

\begin{lstlisting}
You are an expert that is good at judging whether the response to a given query meets the specified evaluator questions.
Your task is to carefully examine the response to determine if it adheres to each requirement outlined in the evaluator questions.

[Query] {query}
[Response] {response}
[Evaluator Question] {question}

For each question, please provide a justification for your evaluation, explaining how the response does or does not satisfy the criteria and a score ('YES' or 'NO') indicating whether the answer satisfies each constraint.

You should only respond in the following JSON format:
{
    "Question 1": {
        "explanation": "",
        "score": "YES" or "NO"
    },
    "Question 2": {
        "explanation": "",
        "score": "YES" or "NO"
    }
}
\end{lstlisting}

\section{Source Parallel Corpora}
\label{app:corpora}

Table~\ref{tab:corpora} lists the 10 publicly available English-Thai parallel corpora forming the $\sim$17.85M-record source pool, after unification into Alpaca \texttt{(instruction, input, output)} format. Phase 1 (Section~\ref{sec:phase1}) cherry-selects from this pool, and Phase 2 augments the selected subset.

\begin{table}[ht]
\centering
\footnotesize
\begin{tabularx}{\columnwidth}{lX}
\toprule
\textbf{Corpus} & \textbf{Type} \\
\midrule
\texttt{bible}         & Religious parallel \citep{christodoulopoulos2015bible} \\
\texttt{paracrawl}     & Web crawl \citep{koehn2024paracrawl} \\
\texttt{elrc}          & Government / news \citep{tiedemann2012opus} \\
\texttt{hplt}          & Web crawl \citep{degibert2024hplt} \\
\texttt{en-th-texts}   & Mixed \citep{kvush_enth} \\
\texttt{opensubtitles} & Film/TV subtitles \citep{lison2016opensubtitles} \\
\texttt{scb\_2020}     & Mixed \citep{lowphansirikul2020scb} \\
\texttt{tatoeba}       & Crowdsourced pairs \citep{tiedemann2020tatoeba} \\
\texttt{wikimedia}     & Encyclopedic \citep{tiedemann2012opus} \\
\texttt{xlent}         & Web-mined \citep{elkishky2021xlent} \\
\bottomrule
\end{tabularx}
\caption{The 10 publicly available English-Thai parallel corpora (all en$\leftrightarrow$th) unified into the RGDC training pool, used by Phase 1 (cherry selection) and Phase 2 (constraint augmentation).}
\label{tab:corpora}
\end{table}

\section{Grounded Dataset Record Example and Quality Audit}
\label{app:record-example}

\noindent\textbf{Record example.} A Grounded record in Alpaca \texttt{(instruction, input, output)} format. \texttt{instruction} combines task header, source, and \texttt{Rules:} block; \texttt{input} is empty; \texttt{output} is the constraint-\mbox{compliant translation.}

\begin{lstlisting}[escapechar=\|]
[instruction]
Task: Translate Thai -> English.
Source:
|{\thaifont จะส่งไปในอีก}| 35 |{\thaifont นาทีค่ะ}|
### Response:

Rules:
- Use a customer service professional style.
- The translation must consist of exactly one sentence.
- Include the number 35 in the output.

[output]
The item will be dispatched in 35 minutes.
\end{lstlisting}

\texttt{[instruction]} and \texttt{[output]} mark the Alpaca fields and are not stored.

\noindent\textbf{Quality of the regenerated targets.} We score 20{,}000 regenerated targets, 5{,}000 per direction and variant, and their source-corpus references with CometKiwi \citep{rei-etal-2022-cometkiwi}, which is independent of our judge. The regenerated targets match or exceed their references on $85.9\%$ of records overall and on more than $80\%$ in each direction (Table~\ref{tab:phase2-audit}). Since reference-free estimation can mildly favor fluent output, we read this as no quality regression from constraint conditioning. The retained constraints are also mostly semantic, with content, linguistic, and stylistic categories at $77\%$ against format and numerical at $23\%$.

\begin{table}[ht]
\centering
\small
\begin{tabular}{lrrrr}
\toprule
\textbf{Slice} & $n$ & $\Delta$\textbf{CK} & \textbf{SE} & \textbf{hyp\,$\geq$\,ref} \\
\midrule
 Overall              & 20{,}000 & $+13.10$ & 0.09 & $85.9\%$ \\
en$\rightarrow$th    & 10{,}000 & $+16.03$ & 0.13 & $90.9\%$ \\
th$\rightarrow$en    & 10{,}000 & $+10.18$ & 0.13 & $80.8\%$ \\
\bottomrule
\end{tabular}
\caption{CometKiwi audit of Phase 2C regeneration. $\Delta$CK is the mean paired difference between the regenerated target and its source-corpus reference, and hyp\,$\geq$\,ref is the share of records where the regenerated target scores at least as high.}
\label{tab:phase2-audit}
\end{table}

\section{Evaluation Prompt Templates}
\label{app:eval-prompts}

Plain and Constrained prompt formats from Section~\ref{sec:eval-data}. The scaffolding is identical across directions, and for the th-en direction, \texttt{Translate English to Thai.} and \texttt{EN:} become \texttt{Translate Thai to English.} and \texttt{TH:}.

\noindent\textbf{Plain.}

\begin{lstlisting}
Translate English to Thai.

EN: <source text>
\end{lstlisting}

\noindent\textbf{Constrained.}

\begin{lstlisting}
Translate English to Thai.
Rules:
- Return only the translated text
- Use a clear, professional tone in Thai
- Keep all numerals in Arabic digits

EN: <source text>
\end{lstlisting}

\section{Models Compared}
\label{app:models-compared}

Table~\ref{tab:models} lists the ChindaMT variants we evaluate and the external baselines used in the main results.

\begin{table}[ht]
\centering
\small
\begin{tabular}{l r}
\toprule
\textbf{Model} & \textbf{Size} \\
\midrule
\multicolumn{2}{l}{\textbf{ChindaMT (ours)}} \\
\quad ChindaMT (Qwen3.5, 4B primary) & 4B, 2B, 0.8B \\
\quad ChindaMT-Qwen3-4B (cross-gen)  & 4B \\
\multicolumn{2}{l}{\textbf{External baselines}} \\
\quad Typhoon-Translate-1.5          & 4B \\
\quad Hunyuan-MT-1.5                 & 7B, 1.8B \\
\quad GemmaX2-28                     & 9B, 2B \\
\quad TranslateGemma                 & 4B \\
\quad MiLMMT-46                      & 4B, 1B \\
\bottomrule
\end{tabular}
\caption{Models compared. TranslateGemma-4B and MiLMMT-46-4B are 4B-only comparators.}
\label{tab:models}
\end{table}

\section{Implementation Hyperparameters}
\label{app:impl-hyperparams}

Three hyperparameter sets cover the pipeline: Phase 2 auxiliary-LLM decoding (Table~\ref{tab:phase2-sampling}), shared fine-tuning across all four variants (Table~\ref{tab:hyperparams}), and test-time inference (below).

\noindent\textbf{Compute.} All fine-tuning ran on two NVIDIA H100s under DeepSpeed ZeRO-2, about 68 wall-clock hours across the four variants (roughly 25 hours per 4B model, 10 for 2B, and 8 for 0.8B), and the cross-family Gemma-3-4B run about 7 hours. Phase 1 took about two and a half days on one H100, mostly IFD scoring of the pool. Phase 2 data generation, the dominant cost, ran on the Qwen3.5-35B-A3B-FP8 vLLM endpoint over roughly two weeks of active generation.

\begin{table}[ht]
\centering
\small
\begin{tabular}{llr}
\toprule
\textbf{Sub-step} & \textbf{Temp.} & \textbf{Max tokens} \\
\midrule
2A. Constraint extraction & 0.7 & 1024 \\
2B. Evaluation questions  & 0   & 512  \\
2C. Constrained responses & 0.7 & 512  \\
2D. Judge evaluation      & 0   & 1024 \\
\bottomrule
\end{tabular}
\caption{Temperature and max-token settings used by the auxiliary LLM at each Phase 2 sub-step. Identical across runs and bases.}
\label{tab:phase2-sampling}
\end{table}

\begin{table}[ht]
\centering
\small
\begin{tabular}{lr}
\toprule
\textbf{Parameter} & \textbf{Value} \\
\midrule
Method               & SFT (full parameters) \\
Epochs               & 1 \\
Learning rate        & $2 \times 10^{-5}$ \\
Scheduler            & inverse-square-root, $1\%$ warmup \\
Weight decay         & 0.01 \\
Optimizer            & AdamW ($\beta_1{=}0.9$, $\beta_2{=}0.999$) \\
Effective batch size & 64 (2 GPUs) \\
Cutoff length        & 1024 \\
Seed                 & 42 \\
\bottomrule
\end{tabular}
\caption{Core ChindaMT-4B fine-tuning hyperparameters. The same recipe is used for all four variants.}
\label{tab:hyperparams}
\end{table}

\noindent\textbf{Test-time inference.} Our fine-tunes and their bases decode with temperature 0.01, top-$p$ 0.7, top-$k$ 20, repetition penalty 1.05, and up to 1024 new tokens, run as batched HuggingFace inference on a single H100. External baselines use per-model settings, temperature 0.2 and top-$p$ 0.9 for GemmaX2 and Typhoon-Translate, 0.7 and 0.6 for HY-MT-1.5, greedy decoding for MiLMMT-46, and up to 512 new tokens for TranslateGemma. The same configurations are used for the LC\% pairwise tables (Section~\ref{sec:eval-protocol}) and the external-metric benchmarks (Section~\ref{sec:external-bench}).

\section{Evaluation Judge Prompt}
\label{app:judge}

Prompt for the pairwise LC judge (Qwen3.6-35B-A3B-FP8) used in every pairwise table of Section~\ref{sec:results}. Output follows the AlpacaEval-v2 \texttt{ranking\_parser} format with a three-tier rubric and a hard penalty for any content beyond the translation. Decoded with temperature $0$, \texttt{enable\_thinking=false}, max\_tokens $1024$.

\begin{lstlisting}
<|im_start|>system
You are an expert evaluator for translation quality. Evaluate based on:

A) Accuracy & Faithfulness (highest priority)
- Translate correctly, accurately, and completely according to the given content.
- Preserve ALL factual information, meaning, numbers, names, and intent.
- Do NOT add, omit, distort, or hallucinate any content.

B) Instruction Compliance & Format Preservation (equally critical)
- STRICTLY follow every instruction and rule in the original prompt.
- Preserve formatting/structure exactly when required (line breaks, punctuation, tags, placeholders, bulleting, spacing, casing, etc.).
- If the instruction requires specific constraints (e.g., "output ONLY the translation" / "no extra words" / "keep proper nouns"), enforce them.

C) Naturalness & Fluency (only after A and B)
- Translation should read naturally in the target language, like a human translation.
- Correct grammar, appropriate register, and smooth phrasing.

HARD PENALTY RULE (very important):
- Judge ONLY the translated text in each model output.
- If a model output contains ANY extra content that is not part of the translation (e.g., explanations, apologies, notes, metadata, quotes of the prompt, commentary, headings, "Here is the translation:", etc.), it MUST receive a significant penalty.
- If the instruction says the output must contain ONLY the translation, then ANY extra words make that output worse, even if the translation itself is good.

TIE-BREAKERS:
1) Fewer instruction/format violations wins.
2) If still tied, the more accurate/complete one wins.
3) If still tied, the more natural/fluently phrased one wins.
<|im_end|>
<|im_start|>user
I will give you the instructions (prompts) given to the models, and the responses of two models. Please rank the models based on which responses would be preferred by humans and above criteria. All inputs and outputs should be python dictionaries.

Here is the prompt:
{
    "instruction": """{instruction}""",
}

Here are the outputs of the models:
[
    {
        "model": "model_1",
        "answer": """{output_1}"""
    },
    {
        "model": "model_2",
        "answer": """{output_2}"""
    }
]

Now please rank the models by the quality of their answers, so that the model with rank 1 has the best output based on the criteria. Then return a list of the model names and ranks, i.e., produce the following output:
[
    {'model': <model-name>, 'rank': <model-rank>},
    {'model': <model-name>, 'rank': <model-rank>}
]

Your response must be a valid Python list and should contain nothing else because we will directly execute it in Python.
Do NOT use markdown code blocks or ```. Return ONLY the raw Python list.
<|im_end|>
\end{lstlisting}

\section{Evaluation Metrics Reference}
\label{app:metrics}

Definitions, strengths, and table pointers for every quality metric used in the paper.

\subsection{Raw win rate (WR\%)}

Share of items where the judge prefers the target over the reference, with ties as half-wins. Given $N$ items with wins $w$, losses $\ell$, draws $d$:
\[\mathrm{WR}\% = (w + 0.5 \cdot d) \,/\, N \cdot 100.\]
WR\% summarizes per-item votes and shares units with LC\%, but inherits length bias because LLM judges prefer verbose responses \citep{saito2023verbosity, dubois2024lcalpacaeval}. Used in Table~\ref{tab:main-direction}, and inside every LC\% comparison.

\subsection{Length-controlled win rate (LC\%)}

The win rate the target would have at matched reference length. AlpacaEval-v2 fits a logistic GLM predicting \texttt{preference} from a standardized length difference $\Delta\ell$ and per-instruction difficulty $d$,
\[P(\text{target wins} \mid \text{item}) = \sigma(\alpha \cdot \tanh(\Delta\ell) + \beta \cdot d + \gamma),\]
and evaluates it at $\Delta\ell = 0$, giving $\mathrm{LC}\% = 100 \cdot \mathbb{E}[P(\text{target wins} \mid \Delta\ell{=}0)]$. LC\% removes the length bias, is stable at $N{=}400$, and agrees with WR\% within 1--2 points. It needs $\sim$400 items per comparison, so Table~\ref{tab:main-direction} reports WR\% per direction. The per-direction cells of Table~\ref{tab:main} are separate fits on each direction, so the Mean is not their average. Standard errors are the AlpacaEval-v2 estimate over the raw per-item votes; cross-judge (Section~\ref{sec:cross-judge}) and external win-rate comparisons use a 1000-draw bootstrap. Used in Tables~\ref{tab:main} and~\ref{tab:own-base}.

\subsection{External translation-quality metrics}

Table~\ref{tab:external} reports four metrics, with the GEMBA prompts in Appendix~\ref{app:metric-prompts}.

\noindent\textbf{CometKiwi} \citep{rei-etal-2022-cometkiwi} is a neural reference-free quality estimator in $[0, 100]$ (\texttt{Unbabel/wmt22-cometkiwi-da}) that rewards neither reference-copying nor rule compliance.

\noindent\textbf{GEMBA-DA} \citep{kocmi-federmann-2023-large} prompts our Qwen3.6-35B judge for a $0$--$100$ score from source, reference, and candidate, making it reference-dependent and prone to LLM biases.

\noindent\textbf{GEMBA-MQM} \citep{kocmi-federmann-2023-gemba} uses the same judge to label MQM error spans by severity, aggregated to a $0$--$100$ score, and is reference-free and diagnostic but coarse.

\noindent\textbf{MetricX-24} \citep{juraska-etal-2024-metricx} is a neural reference-based MQM error score on a native $0$--$25$ scale (lower is better); we report it in this native form and exclude it from the external-metric Mean.

\subsection{Metric alignment with LC\%}

Our LC\% judge (Appendix~\ref{app:judge}) scores pairwise on \textit{accuracy > instruction compliance > fluency} with a hard penalty for extra non-translation content. Table~\ref{tab:metric-alignment} compares LC\% to the external metrics across five axes.

\begin{table}[ht]
\centering
\small
\setlength{\tabcolsep}{4pt}
\begin{tabular}{lccccc}
\toprule
\textbf{Metric} & \textbf{Acc.} & \textbf{IF} & \textbf{Flu.} & \textbf{Pair.} & \textbf{Ref.} \\
\midrule
WR\% / LC\%  & yes & yes        & yes         & yes & no  \\
CometKiwi    & yes & no         & yes         & no  & no  \\
GEMBA-DA     & yes & indirectly & yes         & no  & yes \\
GEMBA-MQM    & yes & no         & via Flu.    & no  & no  \\
\bottomrule
\end{tabular}
\caption{Metric coverage across accuracy (Acc.), instruction-following (IF), fluency (Flu.), pairwise comparison (Pair.), and reference dependence (Ref.). LC\% covers all five, our primary metric. External metrics (Section~\ref{sec:external-bench}) check translation quality alone.}
\label{tab:metric-alignment}
\end{table}

\section{Per-Direction Pairwise Win Rates}
\label{app:per-direction}

Table~\ref{tab:main-direction} reports per-direction win rates by data source. The CoreEval columns break the shared-prompt comparisons of Table~\ref{tab:main} into directions; the external columns apply the same protocol (Section~\ref{sec:eval-protocol}) to FLORES and WMT24++, which carry plain translation only. The external pattern matches Table~\ref{tab:external}: ChindaMT-4B and ChindaMT-2B lead all size-matched FLORES comparisons and most on WMT24++, with the HY-MT news-domain edge clearest against the larger 7B specialist on WMT24++ th-en.

\begin{table*}[t]
\centering
\small
\setlength{\tabcolsep}{4.5pt}
\begin{tabular}{llrrrrrrrr}
\toprule
\multirow{3}{*}{\textbf{Model}} & \multirow{3}{*}{\textbf{Comparator}} & \multicolumn{4}{c}{\textbf{CoreEval}} & \multicolumn{4}{c}{\textbf{External (plain)}} \\
\cmidrule(lr){3-6} \cmidrule(lr){7-10}
 & & \multicolumn{2}{c}{Plain} & \multicolumn{2}{c}{Constrained} & \multicolumn{2}{c}{FLORES} & \multicolumn{2}{c}{WMT24++} \\
\cmidrule(lr){3-4}\cmidrule(lr){5-6}\cmidrule(lr){7-8}\cmidrule(lr){9-10}
 & & en$\rightarrow$th & th$\rightarrow$en & en$\rightarrow$th & th$\rightarrow$en & en$\rightarrow$th & th$\rightarrow$en & en$\rightarrow$th & th$\rightarrow$en \\
\midrule
\multirow{5}{*}{\textbf{ChindaMT-4B}}
  & MiLMMT-46-4B          & 88.8 & 84.5 & 90.5 & 90.0 & 55.3 & 52.5 & 51.4 & 39.9 \\
  & TranslateGemma-4B     & 86.8 & 72.5 & 83.8 & 87.5 & 73.9 & 64.7 & 71.9 & 53.3 \\
  & Typhoon-Translate-4B  & 65.0 & 59.3 & 66.0 & 69.8 & 55.4 & 60.9 & 55.1 & 49.5 \\
  & HY-MT-1.5-7B          & 65.5 & 78.5 & 94.5 & 98.5 & 61.0 & 60.8 & 48.3 & 38.9 \\
  & GemmaX2-28-9B         & 95.0 & 92.5 & 95.5 & 94.8 & 67.5 & 58.2 & 68.0 & 44.8 \\
\cmidrule(l){2-10}
\multirow{3}{*}{\textbf{ChindaMT-2B}}
  & MiLMMT-46-1B          & 88.5 & 87.0 & 92.5 & 93.5 & -- & -- & -- & -- \\
  & HY-MT-1.5-1.8B        & 72.3 & 75.0 & 80.0 & 74.3 & 73.8 & 76.5 & 64.7 & 53.1 \\
  & GemmaX2-28-2B         & 93.5 & 89.3 & 93.5 & 91.3 & 67.9 & 54.7 & 67.8 & 47.1 \\
\cmidrule(l){2-10}
\multirow{3}{*}{\textbf{ChindaMT-0.8B}}
  & MiLMMT-46-1B          & 86.0 & 79.5 & 93.0 & 86.0 & -- & -- & -- & -- \\
  & HY-MT-1.5-1.8B        & 62.5 & 64.5 & 72.0 & 62.0 & 57.0 & 59.7 & 45.1 & 38.0 \\
  & GemmaX2-28-2B         & 93.0 & 85.0 & 93.5 & 86.0 & 50.3 & 38.0 & 49.4 & 31.8 \\
\bottomrule
\end{tabular}
\caption{Per-direction pairwise win rate (WR\%) by data source. CoreEval columns break the shared-prompt comparisons of Table~\ref{tab:main} into directions (Plain and Constrained); external columns apply the same pairwise protocol (Section~\ref{sec:eval-protocol}) to FLORES and WMT24++, which carry plain translation only. Dashes mark comparisons not run, as MiLMMT-46-1B has no external pairing. CoreEval columns report raw WR\% per direction; external bootstrap SE is $[1.3, 1.6]$.}
\label{tab:main-direction}
\end{table*}

\section{External-Benchmark Metric Prompts}
\label{app:metric-prompts}

CometKiwi is a neural model and uses no prompt. For GEMBA-DA and GEMBA-MQM we adapt the prompts of \citet{kocmi-federmann-2023-large} and \citet{kocmi-federmann-2023-gemba} to our Qwen3.6-35B judge. GEMBA-DA keeps the direct-assessment format and returns a $0$--$100$ score from source, reference, and candidate. GEMBA-MQM is zero-shot with six coarse categories, and its labeled errors are aggregated into a $0$--$100$ score by subtracting 1, 5, and 10 per minor, major, and critical error. On a parse failure the judge is re-asked with a one-line format reminder. The two prompts follow, with placeholders in braces.

\noindent\textbf{GEMBA-DA prompt.}
\begin{lstlisting}
You are evaluating the quality of a machine translation.

Source ({src_lang}):
{src}

Reference translation ({tgt_lang}):
{ref}

Candidate translation ({tgt_lang}):
{mt}

Rate the candidate's overall translation quality on a scale from 0 to 100, where:
- 0-30: major meaning errors, poor fluency, unusable
- 30-60: partial meaning preserved, noticeable errors
- 60-80: good translation, minor issues
- 80-100: excellent translation, near-perfect

Consider accuracy, fluency, and faithfulness to the source. The reference is one acceptable translation; equally valid alternatives should not be penalized.

Output only a single integer between 0 and 100, nothing else.
\end{lstlisting}

\noindent\textbf{GEMBA-MQM prompt.}
\begin{lstlisting}
You are performing a Multidimensional Quality Metrics (MQM) evaluation.

Source ({src_lang}):
{src}

Candidate translation ({tgt_lang}):
{mt}

Identify translation errors in the candidate. For each error, output one line in the exact format:
  CATEGORY / SEVERITY / short-description

Categories: Accuracy, Fluency, Terminology, Style, Locale, Other
Severities: Minor, Major, Critical

If there are no errors, output the single line:
  NO_ERRORS

Do not include any text other than the error lines.
\end{lstlisting}

\section{Per-Judge Cross-Judge Numbers}
\label{app:cross-judge-full}

Table~\ref{tab:cross-judge-full} reports the per-judge WR\% underlying the consensus mean in Table~\ref{tab:cross-judge}. Columns Pri, OSS, and Llama are the three open-weight judges, Qwen3.6-35B (primary), GPT-OSS-120B, and Llama-3.3-70B-Instruct, with identical prompts.

\section{Human Evaluation Protocol}
\label{app:human}

Three native Thai raters, one with a translation background, each rated 100 stratified items from the ChindaMT-4B against Typhoon-Translate-1.5-4B comparison, Plain with the baseline template and Constrained with the shared template, 50 Plain and 50 Constrained, A/B-blinded on a 5-point ($+2$ to $-2$) preference scale toward ChindaMT. The win rate collapses this to win/tie/loss by sign (ties as half-wins) and intensity is the mean rating. Two attention-check items per rater were excluded. Results are in Table~\ref{tab:human}.

\noindent\textbf{Rater instructions (translated from Thai).} For each pair, A and B in random order (Constrained items also show a Rules block), raters chose the better translation, judging in priority order accuracy and faithfulness, then rule compliance, then fluency, and scored it 2 (clearly better), 1 (slightly better), or 0 (tie). A few items were attention checks with an obviously incorrect option. Participation was voluntary with the right to withdraw, and only preferences were recorded. After unblinding, preferences are oriented toward ChindaMT for the $+2$ to $-2$ scale in Table~\ref{tab:human}.

\section{Method Lift and Parameter Scaling}
\label{app:scaling}

\enlargethispage{\baselineskip}
Referenced from Section~\ref{sec:own-base}. Table~\ref{tab:own-base} reports two scaling axes on CoreEval, each RGDC fine-tune against its own base (upper block), one step of within-Qwen3.5 parameter halving at fixed recipe (middle block), and the cross-family Gemma-3-4B fine-tune (last row).

\begin{table}[t]
\centering
\footnotesize
\setlength{\tabcolsep}{3pt}
\begin{tabular}{llrr}
\toprule
\textbf{Model} & \textbf{Comparator} & \textbf{Plain} & \textbf{Constr.} \\
\midrule
ChindaMT-4B       & Qwen3.5-4B   & 55.7 & 62.2 \\
ChindaMT-Qwen3-4B & Qwen3-4B     & 71.9 & 69.5 \\
ChindaMT-2B       & Qwen3.5-2B   & 77.1 & 74.2 \\
ChindaMT-0.8B     & Qwen3.5-0.8B & 81.5 & 83.5 \\
\midrule
ChindaMT-4B       & ChindaMT-2B   & 62.6 & 64.2 \\
ChindaMT-2B       & ChindaMT-0.8B & 56.9 & 64.4 \\
\midrule
 Gemma-3-4B FT & Gemma-3-4B & 99.1 & 66.6 \\
\bottomrule
\end{tabular}
\caption{LC\% on CoreEval. Upper block: each ChindaMT variant vs its own base; middle block: ChindaMT variants across parameter halving; last row: the cross-family Gemma-3-4B fine-tune (FT) vs its base. SE $[1.8, 2.4]$, {} $[2.2, 2.4]$, and $0.7/2.3$ by block, the last small because the Plain cell is near ceiling. Under the primary, GPT-OSS-120B, and Llama-3.3-70B judges, the Gemma-3-4B row has Plain/Constrained WR\% of $97.9/67.4$, $96.2/62.4$, and $97.9/67.0$ ($n{=}400$).}
\label{tab:own-base}
\end{table}

\section{External-Metric Ablations at 4B}
\label{app:ablation-external}

Table~\ref{tab:ablation-ext} reports direction-averaged external metrics per RGDC ablation, complementing the LC\% ablations in Table~\ref{tab:ablation-lc}.

\begin{table*}[t]
\centering
\small
\setlength{\tabcolsep}{4pt}
\begin{tabular}{cllrrrrrrrrr}
\toprule
& \multirow{2}{*}{\textbf{Model}} & \multirow{2}{*}{\textbf{Comparator}} & \multicolumn{3}{c}{\textbf{Plain (shared)}} & \multicolumn{3}{c}{\textbf{Plain (own)}} & \multicolumn{3}{c}{\textbf{Constrained}} \\
\cmidrule(lr){4-6} \cmidrule(lr){7-9} \cmidrule(lr){10-12}
& & & Pri & OSS & Llama & Pri & OSS & Llama & Pri & OSS & Llama \\
\midrule
\multirow{11}{*}{\rotatebox[origin=c]{90}{\textsc{CoreEval}}}
  & \multirow{5}{*}{\textbf{ChindaMT-4B}}   & MiLMMT-46-4B            & 86.6 & 86.5 & 86.4 & 55.1 & 47.1          & 50.9 & 90.2 & 89.5 & 89.0 \\
  &                                          & TranslateGemma-4B       & 79.6 & 77.8 & 78.6 & 66.5 & 59.9 & 63.5 & 85.6 & 83.2 & 86.4 \\
  &                                          & Typhoon-Translate-4B    & 62.1 & 55.0 & 54.1 & 59.8 & 52.2 & 56.2 & 67.9 & 66.1 & 62.6 \\
  &                                          & HY-MT-1.5-7B            & 72.0 & 61.6 & 65.5 & 62.1 & 53.0 & 56.6 & 96.5 & 94.8 & 94.8 \\
  &                                          & GemmaX2-28-9B           & 93.8 & 93.0 & 92.2 & 59.4 & 54.1 & 55.6 & 95.1 & 92.8 & 93.1 \\
\cmidrule(l){3-12}
  & \multirow{3}{*}{\textbf{ChindaMT-2B}}   & MiLMMT-46-1B            & 87.8 & 85.8 & 84.8 & 55.6 & 51.9 & 54.9 & 93.0 & 90.8 & 90.2 \\
  &                                          & HY-MT-1.5-1.8B          & 73.6 & 60.0 & 66.1 & 70.4 & 54.6 & 61.4 & 77.1 & 70.9 & 72.1 \\
  &                                          & GemmaX2-28-2B           & 91.4 & 90.3 & 89.1 & 58.4 & 54.0 & 55.6 & 92.4 & 91.0 & 92.4 \\
\cmidrule(l){3-12}
  & \multirow{3}{*}{\textbf{ChindaMT-0.8B}} & MiLMMT-46-1B            & 82.8 & 84.5 & 84.5 & 52.6 & 51.2 & 51.1 & 89.5 & 87.5 & 87.5 \\
  &                                          & HY-MT-1.5-1.8B          & 63.5 & 58.5 & 61.0 & 62.3 & 51.0 & 57.8 & 67.0 & 62.5 & 65.8 \\
  &                                          & GemmaX2-28-2B           & 89.0 & 89.8 & 88.0 & 54.4 & 50.1 & 54.6 & 89.8 & 90.3 & 90.3 \\
\midrule
\multirow{11}{*}{\rotatebox[origin=c]{90}{\textsc{BroadEval}}}
  & \multirow{5}{*}{\textbf{ChindaMT-4B}}   & MiLMMT-46-4B            & ---           & ---           & ---           & 59.4 & 51.6 & 66.6 & ---           & ---           & ---           \\
  &                                          & TranslateGemma-4B       & 81.6 & 81.1 & 77.1 & 68.5 & 64.6 & 69.5 & 74.2 & 73.5 & 73.2 \\
  &                                          & Typhoon-Translate-4B    & 60.0 & 53.2 & 56.8 & 56.4 & 50.9 & 58.9 & 62.8 & 55.9 & 61.5 \\
  &                                          & HY-MT-1.5-7B            & 66.5 & 67.8 & 71.2 & 59.0 & 52.5 & 59.2 & 96.0 & 96.2 & 95.0 \\
  &                                          & GemmaX2-28-9B           & ---           & ---           & ---           & 73.2 & 61.2 & 73.5 & ---           & ---           & ---           \\
\cmidrule(l){3-12}
  & \multirow{3}{*}{\textbf{ChindaMT-2B}}   & MiLMMT-46-1B            & 95.0 & 95.0 & 94.8 & 70.9 & 65.0 & 67.4 & ---           & ---           & ---           \\
  &                                          & HY-MT-1.5-1.8B          & 68.8 & 66.8 & 72.0 & 73.4 & 66.2 & 68.1 & 75.1 & 73.2 & 72.8 \\
  &                                          & GemmaX2-28-2B           & ---           & ---           & ---           & 69.5 & 60.6 & 73.2 & ---           & ---           & ---           \\
\cmidrule(l){3-12}
  & \multirow{3}{*}{\textbf{ChindaMT-0.8B}} & MiLMMT-46-1B            & 93.0 & 93.2 & 92.8 & 52.4 & 49.8          & 57.6 & ---           & ---           & ---           \\
  &                                          & HY-MT-1.5-1.8B          & 51.0 & 51.5 & 59.2 & 51.5 & 51.2 & 55.8 & 60.1 & 63.8 & 64.4 \\
  &                                          & GemmaX2-28-2B           & ---           & ---           & ---           & 52.4 & 45.0          & 59.9 & ---           & ---           & ---           \\
\bottomrule
\end{tabular}
\caption{Per-judge WR\% for the three judges behind the Table~\ref{tab:cross-judge} consensus. Direction-averaged Mean; values above 50 favor ChindaMT. $n{=}400$, bootstrap SE $[0.9, 2.6]$.}
\label{tab:cross-judge-full}
\end{table*}

\noindent\textbf{Cherry selection.} No-cherry slightly edges the full recipe out of domain, with Mean lifts of $0.23$ on FLORES (from $90.65$ to $90.88$) and $0.31$ on WMT24++ (from $87.81$ to $88.12$). The 8.24 Plain CoreEval gain (Section~\ref{sec:ablations}) therefore reflects in-domain alignment, not a uniform quality boost.

\noindent\textbf{All-pass filter.} The full recipe holds the higher Mean on both datasets (FLORES 90.65 vs 90.60, WMT24++ 87.81 vs 87.79), with per-metric deltas scattering near zero without a consistent direction.

\noindent\textbf{Rule-block training.} No-rules loses on every metric in both datasets, with Means dropping $0.92$ on FLORES (from $90.65$ to $89.73$) and $0.72$ on WMT24++ (from $87.81$ to $87.09$). The largest drops come from GEMBA-DA, indicating translation-quality regression rather than narrow instruction-following loss.
\raggedbottom
\begin{table}[ht]
\centering
\footnotesize
\setlength{\tabcolsep}{3pt}
\resizebox{\columnwidth}{!}{%
\begin{tabular}{llrrrr}
\toprule
\textbf{Dataset} & \textbf{Variant} & \textbf{CK} & \textbf{DA} & \textbf{MQM} & \textbf{Mean} \\
\midrule
\multirow{4}{*}{FLORES-200}
  & Full recipe & 84.18 & 89.83 & 97.96 & 90.65          \\
  & No-cherry   & 84.28 & 90.27 & 98.09 & \textbf{90.88} \\
  & No-all-pass & 84.19 & 89.68 & 97.92 & 90.60          \\
  & No-rules    & 83.36 & 88.12 & 97.70 & 89.73          \\
\midrule
\multirow{4}{*}{WMT24++}
  & Full recipe & 80.72 & 85.73 & 97.00 & 87.81          \\
  & No-cherry   & 80.89 & 86.57 & 96.89 & \textbf{88.12} \\
  & No-all-pass & 80.76 & 85.90 & 96.70 & 87.79          \\
  & No-rules    & 80.27 & 84.43 & 96.56 & 87.09          \\
\bottomrule
\end{tabular}}
\caption{External-metric ablations at 4B, direction-averaged; ablations defined in Section~\ref{sec:ablations}. \textbf{Bold} marks the best Mean per dataset.}
\label{tab:ablation-ext}
\end{table}

\section{Qualitative Case Transcript}
\label{app:qualitative-case}

\noindent\begin{minipage}{\linewidth}
Full input and per-model outputs for the rule-violation example in Section~\ref{sec:qualitative}.
\begin{lstlisting}[escapechar=\|]
INSTRUCTION:
Translate Thai to English.
Rules:
- Return only the translated text
- Use concise, natural English
TH: |{\thaifont ท่องเว็บได้แบบส่วนตัวในอุปกรณ์ทุกเครื่อง}|
ChindaMT-4B:
  Browse the web privately on any device.
Qwen3.5-4B:
  Browse privately on any device.
Typhoon-Translate:
  Rules:
  - Return only the translated text
  - Use concise, natural English

  EN: Browse the web privately on every device.
\end{lstlisting}
\end{minipage}

\end{document}